\documentclass[11pt]{article}

\newcommand{\altrowcolors}{\rowcolors{2}{mgray}{white}}

\usepackage[final]{acl}
\usepackage{capt-of}

\usepackage{times}
\usepackage{latexsym}

\usepackage[T1]{fontenc}

\usepackage[utf8]{inputenc}

\usepackage{microtype}

\usepackage{inconsolata}

\usepackage{graphicx}
\usepackage{booktabs, tabularx, multirow, multicol, makecell, colortbl}

\usepackage{enumitem}

\usepackage{bbding}

\usepackage{amsmath, amssymb}
\let\Cross\relax
\usepackage{marvosym}
\usepackage{float}
\usepackage{footmisc}
\definecolor{mblue}{HTML}{D7E3F4}   
\definecolor{mgreen}{HTML}{DDE8D4}  
\definecolor{mgray}{HTML}{EFEDEA}   
\definecolor{mpurple}{HTML}{E7D9F2} 
\definecolor{mred}{HTML}{F1D6D6}    

\definecolor{mheader}{HTML}{E7E4DF}   
\definecolor{mgroup}{HTML}{DDE3E8}    
\definecolor{malt}{HTML}{F6F4F1}      
\definecolor{mbest}{HTML}{E8DFF2}     
\definecolor{msecond}{HTML}{DDEAF6}   

\definecolor{MorandiBlue}{HTML}{5C6F7B}       
\definecolor{MorandiBlueLight}{HTML}{E8ECEF}  
\definecolor{MorandiGreenLight}{HTML}{EDF1EE} 

\usepackage[table]{xcolor}

\title{LaoBench: A Large-Scale Multidimensional Lao Benchmark\\ for Large Language Models}
\newcommand{\equal}{\textsuperscript{$*$}}   
\newcommand{\corr}{\textsuperscript{$\dag$}} 
\newcommand{\plead}{\textsuperscript{$\ddag$}} 

\author{\textbf{Jian Gao}$^{\equal,1}$,\textbf{Richeng Xuan}$^{\equal,2}$, \textbf{Zhaolu Kang}$^{\equal,2,3}$,\\
\textbf{Dingshi Liao}$^{1}$,
\textbf{Wenxin Huang}$^{1}$,
\textbf{Zongmou Huang}$^{1}$,
\textbf{Yangdi Xu}$^{1}$,\\
\textbf{Bowen Qin}$^{2}$,
\textbf{Zheqi He}$^{2}$,
\textbf{Xi Yang}$^{\corr,2}$,
\textbf{Changjin Li}$^{\plead,1}$,
\textbf{Yonghua Lin}$^{\plead,2}$\\
\textsuperscript{1}China-ASEAN Information Harbor Co., Ltd., Nanning, China \\
\textsuperscript{2}Beijing Academy of Artificial Intelligence, Beijing, China \\
 \textsuperscript{3}School of Software \& Microelectronics, Peking University, Beijing, China \\
  }

\begin{document}
\maketitle

\begingroup
  \renewcommand\thefootnote{\fnsymbol{footnote}}
  \footnotetext[1]{Equal contribution.}
  \footnotetext[2]{Corresponding author.}
  \footnotetext[3]{Project lead.}
\endgroup

\begin{table*}[t]
\centering
\scriptsize
\setlength{\tabcolsep}{3pt}
\renewcommand{\arraystretch}{1.15}

\begin{tabular}{lccccccccc}
\toprule
\rowcolor{mheader}
\textbf{Benchmark} & \textbf{Year} & \textbf{SEA Focus} & \textbf{Lao} & \textbf{Native} & \textbf{Knowledge} & \textbf{K12/Exam} & \textbf{Translation} & \textbf{Open Set} & \textbf{Held-out/Black-box} \\
\midrule

\rowcolor{mgroup}
\multicolumn{10}{l}{\textbf{SEA-Region Benchmarks (Holistic / Cultural / Applications)}} \\
\rowcolors{2}{mgray}{white}  

SeaEval~\cite{wang-etal-2024-seaeval}
& 2024 & \Checkmark & \Cross & \Cross & \Checkmark & \Cross & \Cross & \Checkmark & \Checkmark \\

SEA-HELM~\cite{susanto-etal-2025-sea}
& 2025 & \Checkmark & \Cross & \Cross & \Checkmark & \Cross & \Cross & \Checkmark & \Checkmark \\

SeaExam \& SeaBench~\cite{liu-etal-2025-seaexam}
& 2025 & \Checkmark & \Cross & \Checkmark & \Checkmark & \Checkmark & \Cross & \Checkmark & \Cross \\

\midrule

\rowcolor{mgroup}
\multicolumn{10}{l}{\textbf{Language-Specific Benchmarks (Localized / Monolingual)}} \\
\rowcolors{2}{mgray}{white}  

VMLU~\cite{bui-etal-2025-vmlu}
& 2025 & \Cross & \Cross & \Checkmark & \Checkmark & \Cross & \Cross & \Checkmark & \Cross \\

LORAXBENCH~\cite{aji-cohn-2025-loraxbench}
& 2025 & \Checkmark & \Cross & \Checkmark & \Checkmark & \Cross & \Checkmark & \Checkmark & \Cross \\

\midrule

\rowcolor{mgroup}
\multicolumn{10}{l}{\textbf{Broader Low-Resource / Multi-Script Benchmarks (Context)}} \\
\rowcolors{2}{mgray}{white}  

M3Exam~\cite{zhang2023m3exammultilingualmultimodalmultilevel}
& 2023 & \Checkmark & \Cross & \Cross & \Checkmark & \Checkmark & \Cross & \Checkmark & \Cross \\

CIF-Bench~\cite{li-etal-2024-cif}
& 2024 & \Cross & \Cross & \Checkmark & \Cross & \Cross & \Cross & \Checkmark & \Cross \\

MiLiC-Eval~\cite{zhang-etal-2025-milic}
& 2025 & \Cross & \Cross & \Checkmark & \Checkmark & \Cross & \Cross & \Checkmark & \Cross \\

\midrule

\rowcolor{mgroup}
\multicolumn{10}{l}{\textbf{This Work}} \\
\rowcolor{malt}
\textbf{LaoBench (ours)}
& 2026 & \Checkmark & \Checkmark & \Checkmark & \Checkmark & \Checkmark & \Checkmark & \Checkmark & \Checkmark \\

\bottomrule
\end{tabular}

\caption{
Comparison between LaoBench and recent SEA-focused or low-resource evaluation benchmarks.
\textbf{Native} indicates whether the benchmark is originally constructed in the target language(s) rather than translated from English.
\textbf{Open Set} indicates whether a publicly released evaluation set is available for reproducible research.
\textbf{Held-out / Black-box} indicates whether the benchmark provides hidden test evaluation (e.g., held-out test sets or official black-box services) to mitigate contamination and leaderboard overfitting.
}
\label{tab:benchmark_comparison}
\vspace{-6pt}
\end{table*}

\begin{abstract}  
  
The rapid advancement of large language models (LLMs) has not been matched by their evaluation in low-resource languages, especially Southeast Asian languages like Lao. To fill this gap, we introduce \textbf{LaoBench}, the first large-scale, high-quality, and multidimensional benchmark for assessing LLM language understanding and reasoning in Lao. LaoBench contains \textbf{17,000+} expert-curated samples across three dimensions: culturally grounded knowledge application, curriculum-aligned K12 education, and bilingual translation among Lao, Chinese, and English. It includes open-source and held-out subsets, where the held-out portion enables secure black-box evaluation via a controlled service to improve fairness and data security. We construct LaoBench with a hybrid pipeline that combines expert authoring with agent-assisted verification, ensuring linguistic accuracy, cultural relevance, and educational validity. We evaluate diverse state-of-the-art open-source and closed-source LLMs, and find that even strong multilingual models lag behind human experts, particularly in culturally grounded reasoning and translation fidelity. We hope LaoBench will catalyze research on Lao and other underrepresented Southeast Asian languages for more inclusive multilingual evaluation. To facilitate reproducibility, we release LaoBench at \url{https://huggingface.co/datasets/BAAI/LaoBench}.

\end{abstract}

\section{Introduction}

Large language models (LLMs) have achieved strong performance on reasoning, dialogue, and translation~\cite{openai2024gpt4technicalreport}, yet evaluation remains heavily skewed toward high-resource languages. This gap is particularly severe in Southeast Asia, where linguistic diversity is high but benchmark coverage is limited~\cite{FLORES,adelani2021masakhanernamedentityrecognition}. While evaluation resources exist for Thai, Vietnamese, and Indonesian~\cite{yang-etal-2022-cino,xu2025cmhgdatasetbenchmarkheadline,zhang2025milic,raja2025parallelcorporamachinetranslation}, Lao is still largely absent from large-scale multilingual benchmarks (Table~\ref{tab:benchmark_comparison}).

Existing SEA-focused benchmarks also have limitations beyond language coverage. Many emphasize high-level multilingual reasoning or rely on translation from English, which often misses Lao-specific cultural grounding and linguistic properties. Lao’s \textit{scriptio continua} writing system introduces tokenization ambiguity and can distort generation and translation metrics. Curriculum-aligned education evaluation is under-explored, despite being central to native proficiency. Finally, publicly released benchmarks are increasingly vulnerable to contamination and leaderboard overfitting, yet Lao lacks a held-out black-box evaluation service for fair and sustainable comparison. These gaps motivate LaoBench as a native, multidimensional, and contamination-resistant evaluation suite for Lao.

Lao is the official language of Laos and is used by millions of speakers, yet it remains low-resource in NLP due to limited digitized corpora and scarce labeled data. Beyond data scarcity, Lao poses challenges for LLMs, including ambiguous tokenization under \textit{scriptio continua}, a complex tonal system, and frequent Pali--Sanskrit loanwords. Existing Lao NLP resources are mostly task-specific, such as morphological analysis~\cite{eskander-etal-2019-unsupervised} and limited translation datasets~\cite{10.1145/3610404,geigle2024babelimagenetmassivelymultilingualevaluation}, which are insufficient for evaluating general-purpose LLMs. Secure evaluation is also increasingly important as public benchmarks face leakage and overfitting risks, motivating held-out test sets and black-box services~\cite{deng-etal-2024-investigating}.

To address these gaps, we introduce \textbf{LaoBench}, the first large-scale, high-quality, and multidimensional benchmark for evaluating LLMs in Lao. LaoBench contains \textbf{17,000+} instances spanning three dimensions: 
(1) \textit{Knowledge Application} grounded in Lao society, culture, politics, history, and science; 
(2) \textit{K12 Foundational Education} aligned with Lao’s national curriculum; and 
(3) \textit{Bilingual Translation} among Lao, Chinese, and English. 

LaoBench includes three subsets: \textbf{Lao-7k} (7,000 open-source MCQs), \textbf{Lao-10k} (10,000+ held-out MCQs for black-box evaluation), and \textbf{Lao-500} (500 open-ended prompts selected using a BenchBuilder-inspired pipeline~\cite{li2025from}). We construct LaoBench with a rigorous hybrid pipeline combining expert authoring and agent-assisted verification (e.g., duplicate detection, semantic consistency checks, context-independence filtering, and sensitivity screening).

We benchmark diverse state-of-the-art open-source and closed-source LLMs on LaoBench and find a substantial gap to human experts, especially in culturally grounded reasoning and translation fidelity. LaoBench provides a standardized and secure evaluation suite to support reliable comparison and future research on Lao and other underrepresented Southeast Asian languages.

In summary, our contributions are:

\begin{itemize}[leftmargin=*]
    \vspace{-7pt}
    \item \textbf{First multidimensional Lao benchmark.} We introduce \textbf{LaoBench}, the first large-scale benchmark for evaluating LLMs in Lao across (i) culturally grounded knowledge application, (ii) curriculum-aligned K12 education, and (iii) trilingual translation (Lao--Chinese--English). LaoBench contains 17,000+ expert-curated instances.
    \vspace{-6pt}
    \item \textbf{Contamination-resistant held-out evaluation.} 
     We construct a held-out subset of 10,000+ MCQs and design a black-box evaluation protocol to mitigate test leakage and leaderboard overfitting. We will deploy the protocol as an online evaluation service upon publication.
    \vspace{-6pt}
    \item \textbf{Rigorous hybrid construction pipeline.} We develop a scalable pipeline combining expert authoring with agent-assisted verification, including duplicate detection, semantic consistency checks, context-independence filtering, and sensitivity screening.
   \vspace{-6pt}
    \item \textbf{Comprehensive benchmarking and key findings.} We evaluate diverse state-of-the-art open-source and closed-source LLMs on LaoBench and reveal persistent gaps to human experts, especially in culturally grounded reasoning and high-fidelity translation.
\end{itemize}

\section{LaoBench}
\begin{figure*}[!t]
    \centering
    \includegraphics[width=0.93\linewidth]{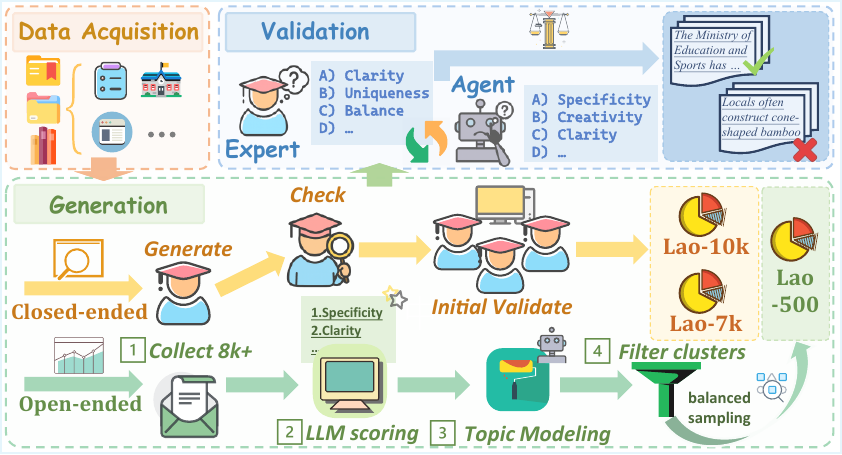}
    \vspace{-3pt}
    \caption{Overview of the LaoBench construction pipeline. We collect raw materials from authoritative Lao sources, construct closed-form multiple-choice questions and open-ended prompts using different strategies, and perform multi-stage validation with expert review and agent-assisted verification to ensure linguistic correctness, cultural relevance, and educational validity.}
    \label{fig:pipeline}
    \vspace{-6pt}
\end{figure*}

LaoBench is a large-scale, multidimensional benchmark designed to systematically evaluate large language models (LLMs) in Lao, a low-resource Southeast Asian language.
LaoBench provides comprehensive evaluation coverage across three core dimensions: \textbf{Knowledge Application}, \textbf{K12 Foundational Education}, and \textbf{Bilingual Translation} among Lao, Chinese, and English.

\subsection{Dataset Overview and Subsets}
LaoBench contains more than \textbf{17,000} carefully curated instances. To support both transparent research usage and secure benchmarking, the benchmark is organized into three complementary subsets, as summarized in Table~\ref{tab:laobench_summary}.
\vspace{-6pt}

\begin{table}[!t]
\centering
\small
\setlength{\tabcolsep}{6pt}
\renewcommand{\arraystretch}{1.15}
\rowcolors{2}{mgray}{white}
\begin{tabular}{lcc}
\toprule
\rowcolor{mheader}
\textbf{Subset} & \textbf{Size} & \textbf{Type} \\
\midrule
\textbf{Lao-7k}  & 7,000     & Multiple-choice, open-source \\
\textbf{Lao-10k} & 10,000+ & Multiple-choice, closed-source \\
\textbf{Lao-500} & 500       & Open-ended prompts, open-source \\
\bottomrule
\end{tabular}
\vspace{-6pt}
\caption{Summary of LaoBench dataset subsets.}
\label{tab:laobench_summary}
\vspace{-10pt}
\end{table}

\begin{figure*}[t]
    \centering
    \includegraphics[width=0.93\linewidth]{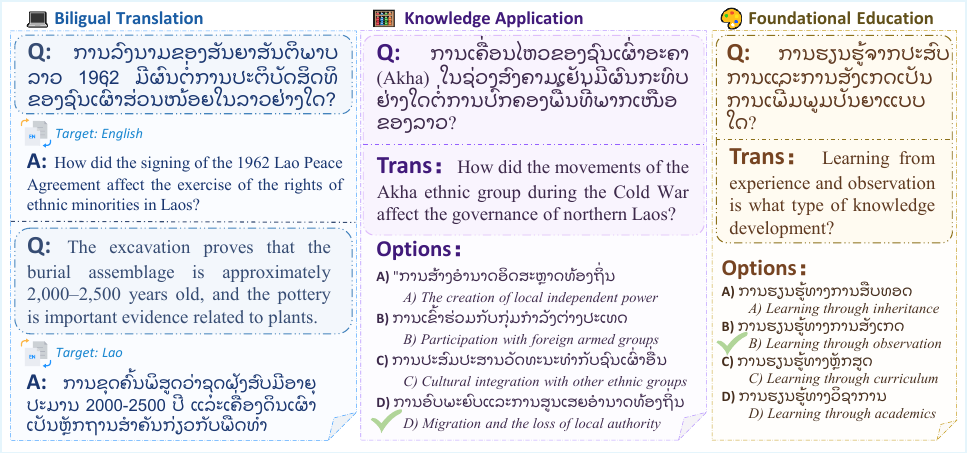}
    \vspace{-6pt}
    \caption{Example cases from LaoBench, illustrating the three task types: Knowledge Application, K12 Education, and Bilingual Translation.}
    \label{fig:case}
    \vspace{-6pt}
\end{figure*}

\paragraph{Lao-7k (open-source multiple-choice).}
Lao-7k contains 7,000 expert-written multiple-choice questions released publicly to enable reproducible benchmarking. These questions span all three evaluation dimensions and are designed to assess both factual understanding and reasoning ability in Lao.
\vspace{-6pt}
\paragraph{Lao-10k (closed-source multiple-choice).}
Lao-10k contains over 10,000 additional multiple-choice questions reserved for black-box evaluation via a controlled black-box service.
We keep this subset hidden and return only aggregated scores to mitigate test leakage and leaderboard overfitting.
The full black-box evaluation protocol is described in Appendix~\ref{sec:appendix.blackbox_protocol}.
\vspace{-2pt}
\paragraph{Lao-500 (open-source open-ended).}
Lao-500 is a set of 500 open-ended prompts intended for evaluating long-form generation and open-domain reasoning in Lao. These prompts are automatically selected from a large candidate pool using a pipeline inspired by BenchBuilder~\cite{li2025from}, which supports continuous benchmark expansion while maintaining diversity and quality.
\vspace{-2pt}
\paragraph{Design rationale of the three subsets.}
We design LaoBench as a combination of open-source and closed-source subsets to balance transparency, fairness, and long-term benchmark reliability.

\subsection{Task Categories and Formats}
Each LaoBench instance belongs to one of the following three categories, designed to capture complementary aspects of Lao language competence:
\vspace{-17pt}
\paragraph{Knowledge Application.}
This category evaluates domain knowledge grounded in Lao society and culture, covering subdomains. Questions are designed to require contextual reasoning and culturally specific knowledge rather than simple pattern matching.
\vspace{-6pt}
\paragraph{K12 Foundational Education.}
This category aligns with Lao’s national K12 curriculum and evaluates foundational knowledge and reasoning skills. Items emphasize educational validity and reflect real classroom-style knowledge requirements.
\vspace{-6pt}
\paragraph{Bilingual Translation.}
This category evaluates translation capabilities among Lao, Chinese, and English in practical domains. Each instance contains a source sentence and a professionally written reference translation. We report corpus-level BLEU under a standardized SacreBLEU configuration with Lao-aware tokenization, and provide full evaluation details in Appendix~\ref{sec:appendix.translation_eval}.

\subsection{Construction Pipeline}
LaoBench is constructed through a three-stage pipeline consisting of \textbf{raw data acquisition}, \textbf{dataset construction}, and \textbf{validation and quality assurance}, as illustrated in Figure~\ref{fig:pipeline}. The pipeline integrates expert human curation with agent-assisted verification to ensure both high quality and scalability.
\vspace{-6pt}

\paragraph{Raw Data Acquisition.}
To ensure that LaoBench reflects realistic language use in Laos, we collect materials from diverse authoritative sources, including K12 textbooks and curriculum guidelines, government and legal documents, encyclopedic and educational publications, as well as culturally grounded articles and local knowledge resources.
These sources cover both formal and informal registers, enabling the benchmark to evaluate not only factual recall but also culturally grounded reasoning and practical language understanding.
By grounding questions in authentic Lao contexts, LaoBench reduces the risk of constructing synthetic or overly translation-based evaluation data, which is a common limitation in low-resource benchmarks.
\vspace{-6pt}
\paragraph{Dataset Construction.}
We adopt two complementary construction strategies depending on the data format. For Lao-7k and Lao-10k, expert linguists and domain specialists create multiple-choice questions by selecting key knowledge points, writing question stems in Lao, and designing plausible distractors with one correct option. Difficulty is calibrated through iterative refinement to reduce ambiguity and ensure meaningful reasoning requirements. For Lao-500, we begin with a large candidate prompt pool derived from the same sources and apply an automated selection procedure inspired by BenchBuilder~\cite{li2025from}. Candidate prompts are scored by an LLM annotator on quality dimensions such as specificity, clarity, domain depth, and creativity. We further apply topic-based clustering to promote diversity, discard low-quality or redundant clusters, and sample evenly from the remaining clusters to form a balanced set of 500 open-ended prompts.
\vspace{-6pt}
\paragraph{Validation and Quality Assurance.}
All items undergo multi-stage validation combining expert review and agent-assisted verification. Human experts verify factual correctness, linguistic fluency, cultural appropriateness, and educational validity. For multiple-choice questions, experts also assess distractor plausibility and remove ambiguous or underspecified items. For translation instances, experts verify semantic alignment between source and reference translations and correct unnatural phrasing. Automated agents further support quality control by detecting duplicates and near-duplicates, checking semantic consistency, and screening potentially sensitive or harmful content. Items failing any validation checks are revised or removed through iterative refinement. 

\begin{figure}[!t]
    \centering
    \includegraphics[width=0.9\linewidth]{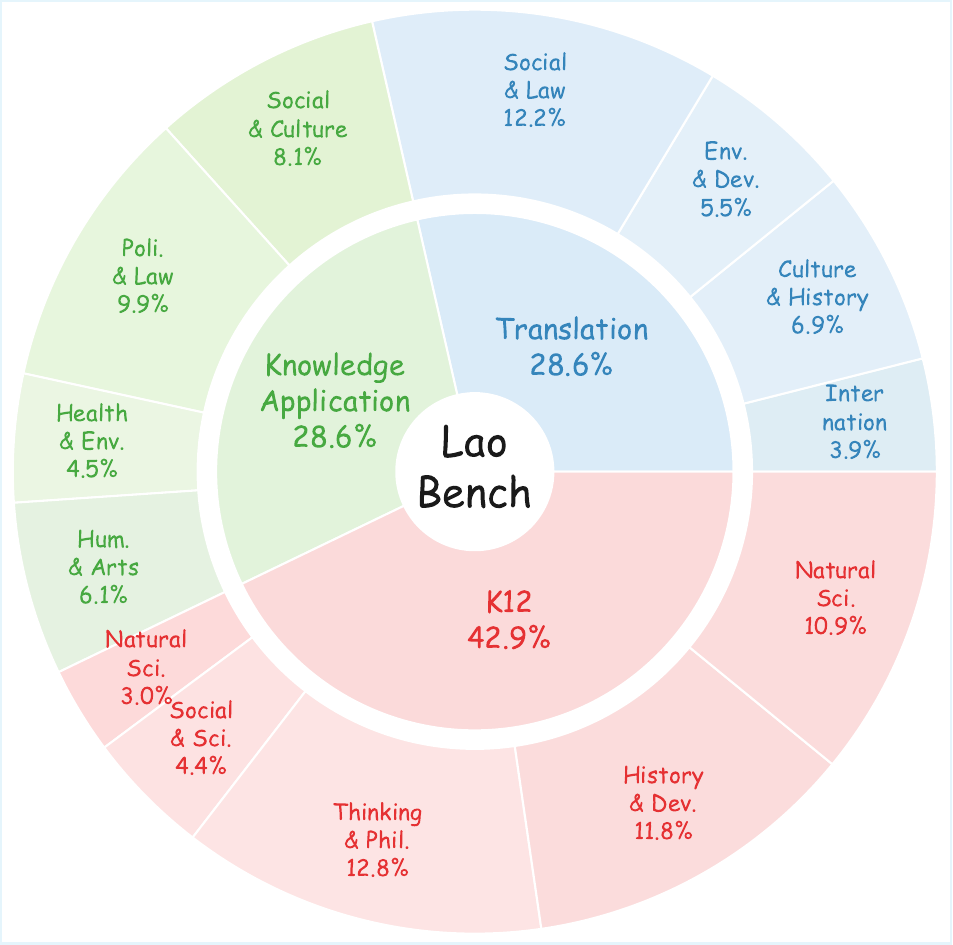}
    \vspace{-3pt}
    \caption{Distribution of Lao-7k samples across the three main categories—Knowledge Application, K12 Education, and Translation—and their subdomains.}
    \label{fig:pie_chart}
    \vspace{-8pt}
\end{figure}

\subsection{Data Statistics}
LaoBench contains more than 17,000 instances across three primary categories: Knowledge Application, K12 Foundational Education, and Bilingual Translation. Each category is further organized into subdomains. The overall distribution of samples is shown in Figure~\ref{fig:pie_chart}. Further details are shown in Appendix~\ref{sec:appendix.detail_stats}. 

\begin{table*}[t]
\centering
\scriptsize
\setlength{\tabcolsep}{1.7pt} 
\renewcommand{\arraystretch}{1.15}

\begin{tabular}{lccccccccccccc}
\toprule
\rowcolor{mheader}
\multirow{2}{*}{\textbf{Model}} 
& \multicolumn{5}{c}{\textbf{K12 Accuracy (\%)}} 
& \multicolumn{4}{c}{\textbf{Translation BLEU}} 
& \multicolumn{4}{c}{\textbf{Knowledge Application Accuracy (\%)}} \\
\cmidrule(lr){2-6} \cmidrule(lr){7-10} \cmidrule(lr){11-14}
& \makecell{Nat.\\Sci.} & \makecell{Soc.\\Sci.} & \makecell{Think.\\\& Phil.} & \makecell{Hum.\\\& Arts} & \makecell{Health\\\& Env.} 
& \makecell{Soc.\\\& Law} & \makecell{Cult.\\\& Hist.} & \makecell{Inter.\\Aff.} & \makecell{Env.\\\& Dev.} 
& \makecell{Pol.\\\& Law} & \makecell{Soc.\\\& Cult.} & \makecell{Hist.\\\& Dev.} & \makecell{Nat.\\Sci.} \\
\midrule

\rowcolor{mgroup}
\multicolumn{14}{l}{\textbf{Blind Evaluation}} \\
Random Choice  
& 25.00 & 25.00 & 25.00 & 25.00 & 25.00 
& -- & -- & -- & -- 
& 25.00 & 25.00 & 25.00 & 25.00 \\
\midrule

\rowcolor{mgroup}
\multicolumn{14}{l}{\textbf{Open-Source Models}} \\
Qwen3-Next-80B-A3B-Instruct 
& 76.15 & 69.94 & 68.32 & 91.15 & 93.43 
& 16.03 & 23.94 & 38.48 & 22.95 
& 68.78 & 67.67 & 60.00 & 55.73 \\

\rowcolor{malt}
Qwen3-235B-A22B 
& 82.75 & 76.48 & 77.09 & 93.11 & 97.65 
& 20.39 & 27.15 & 36.64 & 28.01 
& 71.51 & 71.55 & 62.35 & 57.64 \\

Qwen3-235B-A22B-Instruct-2507 
& 86.45 & 76.36 & 74.74 & 94.75 & 98.59 
& 21.81 & 29.02 & \colorbox{mbest}{41.47} & 29.14 
& 74.82 & 73.85 & 59.53 & 61.46 \\

\rowcolor{malt}
DeepSeek-V3.2-Exp (Thinking) 
& 84.69 & 79.19 & 75.39 & 91.48 & 94.84 
& 20.57 & 28.29 & 34.96 & 27.29 
& 70.94 & 72.32 & 63.76 & 69.43 \\

DeepSeek-V3.2-Exp (Non-Thinking) 
& 79.96 & 72.48 & 68.06 & 91.80 & 95.31 
& 22.52 & 29.77 & 37.16 & 29.49 
& 67.63 & 70.32 & 64.47 & 58.28 \\

\rowcolor{malt}
Ministral-8B-Instruct-2410 
& 23.52 & 30.18 & 25.79 & 34.75 & 27.23 
& 0.83 & 2.09 & 8.61 & 2.09 
& 22.16 & 26.53 & 24.00 & 23.89 \\

Ling-mini-2.0 
& 35.16 & 35.15 & 34.55 & 39.34 & 40.38 
& 0.69 & 2.56 & 9.82 & 2.11 
& 27.91 & 32.33 & 31.76 & 28.98 \\

\midrule

\rowcolor{mgroup}
\multicolumn{14}{l}{\textbf{Closed-Source Models}} \\
GPT-5-High 
& \colorbox{mbest}{90.03} & \colorbox{mbest}{82.91} & 80.76 & 95.08 & \colorbox{mbest}{99.53} 
& 20.96 & 28.52 & 38.59 & 26.91 
& \colorbox{mbest}{79.42} & 77.74 & 66.59 & \colorbox{mbest}{75.80} \\

\rowcolor{malt}
Qwen3-Max 
& 87.35 & 77.82 & 76.31 & 94.75 & 97.65 
& 21.70 & 30.07 & 39.71 & 28.87 
& 75.11 & 76.33 & 61.41 & 63.38 \\

Qwen3-Plus (Non-Thinking) 
& 86.00 & 76.61 & 74.08 & 95.41 & 99.06 
& 21.69 & 28.83 & 40.95 & 28.91 
& 75.54 & 74.73 & 59.29 & 60.51 \\

\rowcolor{malt}
Qwen3-Plus (Thinking) 
& 84.88 & 76.85 & 74.48 & 95.41 & 99.06 
& 20.32 & 28.42 & 39.46 & 28.25 
& 75.83 & 73.85 & 60.47 & 60.51 \\

Gemini-2.5-Pro 
& 88.69 & 82.79 & \colorbox{mbest}{82.20} & 95.08 & 99.06 
& \colorbox{mbest}{26.22} & \colorbox{mbest}{34.31} & 40.71 & \colorbox{mbest}{33.68} 
& 77.55 & \colorbox{mbest}{79.51} & 67.29 & 70.38 \\

\rowcolor{malt}
Claude-Sonnet-4.5-20250929-thinking 
& 89.36 & 80.85 & 79.71 & 95.41 & 98.59 
& 22.76 & 29.96 & 37.50 & 28.41 
& 77.12 & 77.21 & 63.29 & 69.11 \\

Claude-Opus-4.1-20250805 
& 89.03 & 80.00 & 78.27 & \colorbox{mbest}{95.74} & 96.71 
& 24.78 & 32.08 & 38.52 & 32.38 
& 77.84 & 78.80 & \colorbox{mbest}{68.47} & 68.47 \\

\midrule
\rowcolor{mgroup}
\multicolumn{14}{l}{\textbf{Human Evaluation}} \\
Human Experts
& 98.34 & 98.14 & 97.29 & 98.93 & 99.92 
& -- & -- & -- & -- 
& 99.21 & 98.78 & 98.97 & 97.98 \\
\bottomrule
\end{tabular}

\caption{
Detailed performance of evaluated models on Lao-7k.
We report accuracy (\%) for K12 Education and Knowledge Application, and BLEU for Translation.\colorbox{mbest}{Highlighted} cells indicate the best result in each column.
Dimension-level averages for the three core dimensions are reported in Figure~\ref{fig:model_performance_overall}.
}
\vspace{-6pt}
\label{tab:model_performance}
\vspace{-6pt}
\end{table*}

\section{Experiment}

We evaluate LLMs on LaoBench using two complementary protocols.
First, we perform closed-form multiple-choice evaluation on \textbf{Lao-7k}, enabling transparent and reproducible benchmarking with objective metrics.
Second, we conduct open-ended generation evaluation on \textbf{Lao-500} using an Arena-style pairwise comparison framework.
Together, these evaluations provide a holistic view of LLM capabilities in Lao.

\subsection{Experimental Setup}

\paragraph{Models.}
We select representative SOTA closed-source models and diverse open-source families covering different scales and instruction styles
Open-source models include Qwen3-Next-80B-A3B-Instruct~\cite{qwen2.5-1m}, Qwen3-235B-A22B~\cite{qwen3technicalreport}, Qwen3-235B-A22B-Instruct-2507, DeepSeek-V3.2-Exp~\cite{deepseekai2024deepseekv32} (Thinking and Non-Thinking), Ministral-8B-Instruct-2410~\cite{mistralai2024ministral8b}, and Ling-mini-2.0~\cite{Ling-mini-2.0}.
Closed-source models include GPT-5-High~\cite{gpt5}, Qwen3-Max, Qwen3-Plus (Thinking/Non-Thinking), Gemini-2.5-Pro~\cite{comanici2025gemini25pushingfrontier}, Claude-Sonnet-4.5-20250929-thinking~\cite{TheC3}, and Claude-Opus-4.1-20250805.
All API evaluations were conducted during Oct.--Dec. 2025 with deterministic decoding (temperature $=0$ where supported).

\begin{figure*}[!t]
    \centering
    \includegraphics[width=1\linewidth]{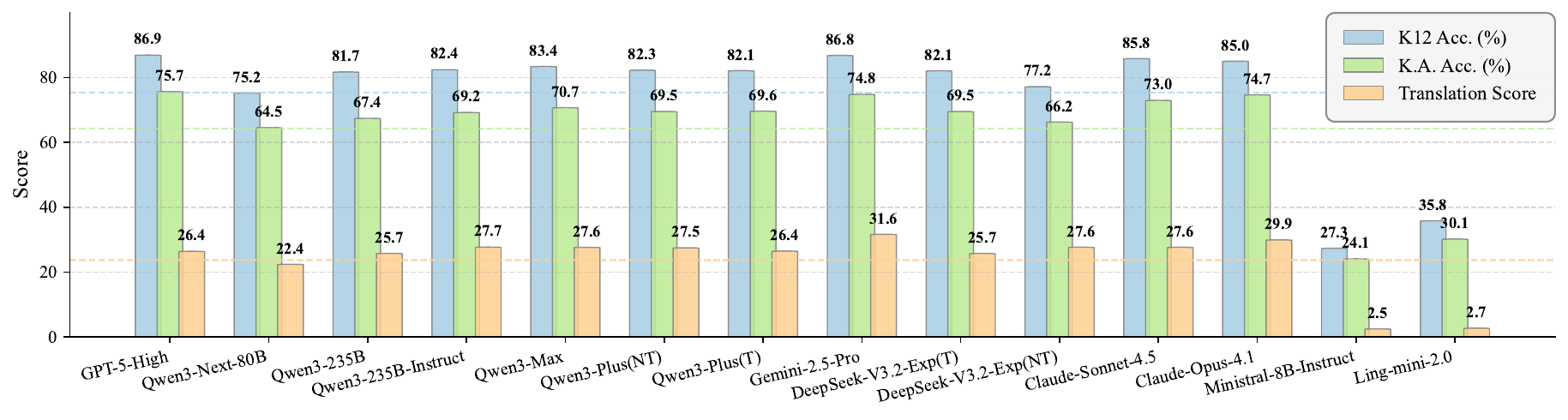}
    \vspace{-6pt}
    \caption{
Overall \textbf{dimension-level averages} (\textit{K12 Avg / Translation Avg / Knowledge Avg}) of evaluated models on Lao-7k.
Each score is averaged over its corresponding subdomains in Table~\ref{tab:model_performance}.
}
\vspace{-6pt}
    \label{fig:model_performance_overall}
\end{figure*}

\paragraph{Prompting and inference.}
All models are evaluated in a \textbf{zero-shot} setting without fine-tuning on LaoBench.
For multiple-choice questions, we present the question stem and four options in Lao and require models to output exactly one option.
For models that support explicit reasoning, we additionally evaluate chain-of-thought (CoT) style prompting (denoted as \emph{Thinking}) and compare against direct-answer prompting (\emph{Non-Thinking}).
For translation evaluation, we use a fixed SacreBLEU configuration with Lao-aware tokenization and additionally report chrF++\cite{popovic-2017-chrf} in Appendix~\ref{sec:appendix.translation_eval}.
\vspace{-4pt}
\paragraph{Output normalization.}
For multiple-choice evaluation, we apply a unified post-processing rule that maps model outputs to a single option label (A/B/C/D). If a model produces multiple labels, non-option text, or an unparseable answer, it is counted as incorrect. 


\subsection{Closed-Form Evaluation on Lao-7k}

We first conduct closed-form evaluation on \textbf{Lao-7k}.
This subset enables transparent and reproducible benchmarking with objective metrics.
\vspace{-6pt}
\paragraph{Metrics.}
For K12 Education and Knowledge Application multiple-choice questions, we report \textbf{Accuracy}.
For Translation tasks, we compute corpus-level \textbf{BLEU} against expert-written references under a standardized SacreBLEU configuration.
Since Lao is written in a scriptio continua style without explicit word boundaries, BLEU can be sensitive to segmentation;
we therefore apply Lao-aware tokenization using LaoNLP before scoring, and additionally report chrF++ in Appendix~\ref{sec:appendix.translation_eval}.
\vspace{-6pt}
\paragraph{Overall results.}
Table~\ref{tab:model_performance} reports detailed results across subdomains for all evaluated models.
A clear gap emerges between open-source and closed-source models: closed-source systems consistently dominate across all dimensions, while open-source models show larger variance.
Among closed-source models, GPT-5-High achieves the strongest overall accuracy on K12 and Knowledge Application, while Gemini-2.5-Pro leads in Translation BLEU, indicating stronger multilingual generation fidelity.
Among open-source models, Qwen3-235B-A22B and DeepSeek-V3.2 are the most competitive, whereas smaller models significantly lag behind in all categories.

\vspace{-2pt}
\paragraph{Fine-grained analysis by subdomain.}
Figure~\ref{fig:model_performance_overall} summarizes model performance aggregated by the three core dimensions.
Across nearly all systems, \textbf{K12 Education} is generally easier than \textbf{Knowledge Application}.
For instance, many strong models exceed 90\% accuracy on K12 subdomains such as Health \& Environment and Humanities \& Arts (Table~\ref{tab:model_performance}), suggesting that structured curriculum-aligned content is relatively accessible to multilingual LLMs.
In contrast, Knowledge Application subdomains yield notably lower accuracy, reflecting the need for culturally grounded and domain-specific reasoning.
Even GPT-5-High shows a substantial drop from K12 to Knowledge Application accuracy (Table~\ref{tab:model_performance}), highlighting the intrinsic difficulty of culturally grounded Lao reasoning.
\begin{figure}[!t]
    \centering
    \includegraphics[width=1\linewidth]{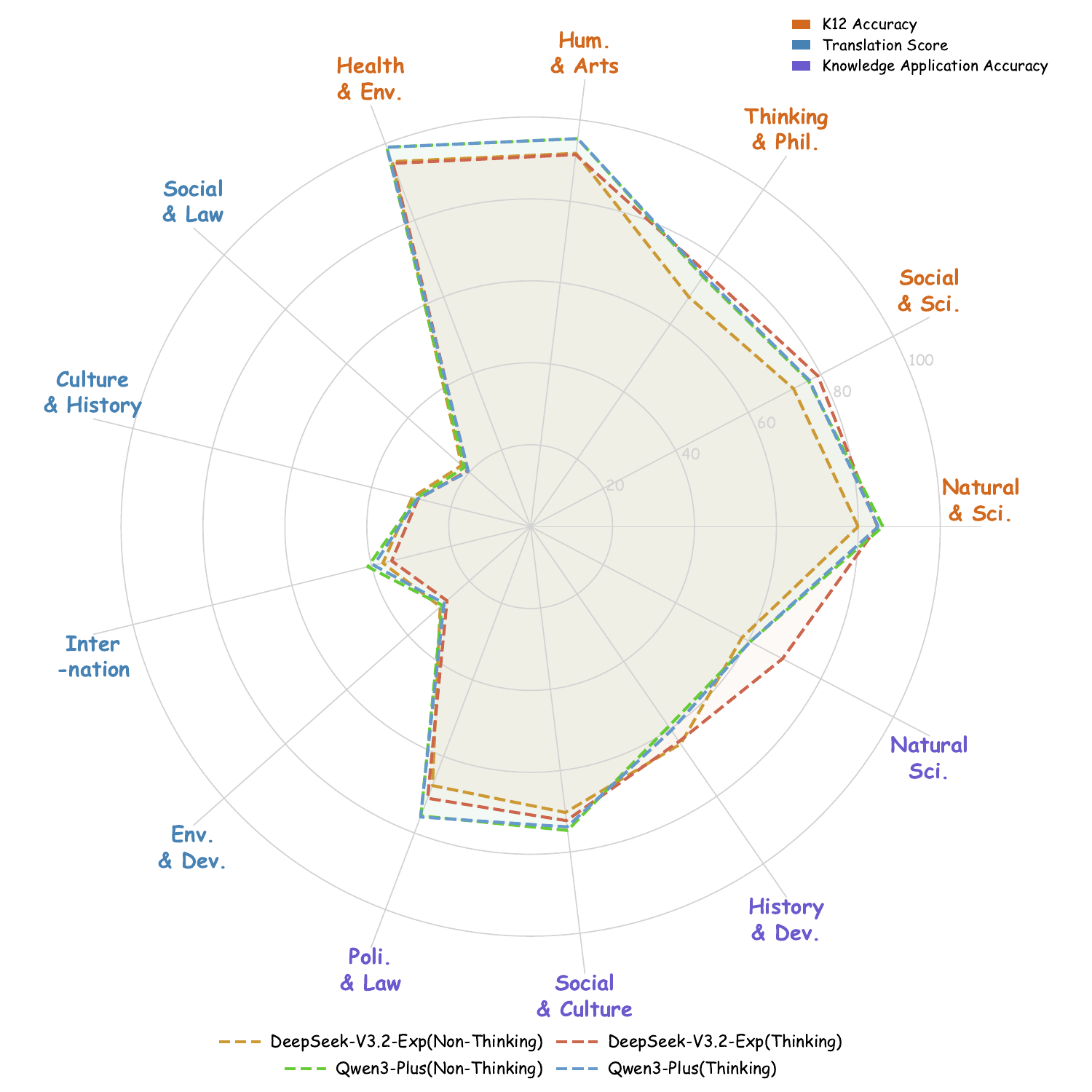}
    \vspace{-6pt}
    \caption{Radar chart comparing performance of models with and without Chain-of-Thought (CoT) prompting on Lao-7k.}
    \vspace{-6pt}
    \label{fig:thinking_comparison}
\end{figure}
\vspace{-3pt}
\paragraph{Translation performance remains limited.}
Translation BLEU scores are modest for all models.
Even the strongest closed-source systems only achieve mid-30s BLEU in the best subdomains, indicating persistent challenges in translation fidelity and fluency.
We also observe topic-dependent differences: Culture \& History and Social \& Law tend to be harder because they contain culturally specific expressions and formal terminology that require precise lexical choice.
These results suggest that translation involving Lao remains challenging even for state-of-the-art multilingual models.


\vspace{-6pt}
\paragraph{Effect of Chain-of-Thought prompting.}
Figure~\ref{fig:thinking_comparison} highlights the impact of Chain-of-Thought (CoT) prompting by comparing Thinking and Non-Thinking variants.
We observe consistent improvements from CoT prompting, especially on subdomains requiring multi-step reasoning and culturally grounded Knowledge Application questions.
However, improvements are smaller on more factual or formulaic K12 subdomains, suggesting that CoT primarily benefits complex reasoning rather than straightforward recall.
\vspace{-6pt}
\paragraph{Gap to human performance.}
Human experts achieve near-perfect accuracy across all K12 and Knowledge Application subdomains establishing a strong upper bound.
The persistent gap between human performance and state-of-the-art models, especially on Knowledge Application and Translation, indicates that LaoBench remains challenging and provides meaningful headroom for future research.

Overall, LaoBench reveals a consistent performance drop from curriculum-aligned knowledge to culturally grounded reasoning and bilingual translation, indicating that Lao remains a challenging low-resource setting for modern LLMs.


\subsection{Open-Ended Arena-Style Evaluation on Lao-500}

\begin{figure}[!t]
    \centering
    \includegraphics[width=\linewidth]{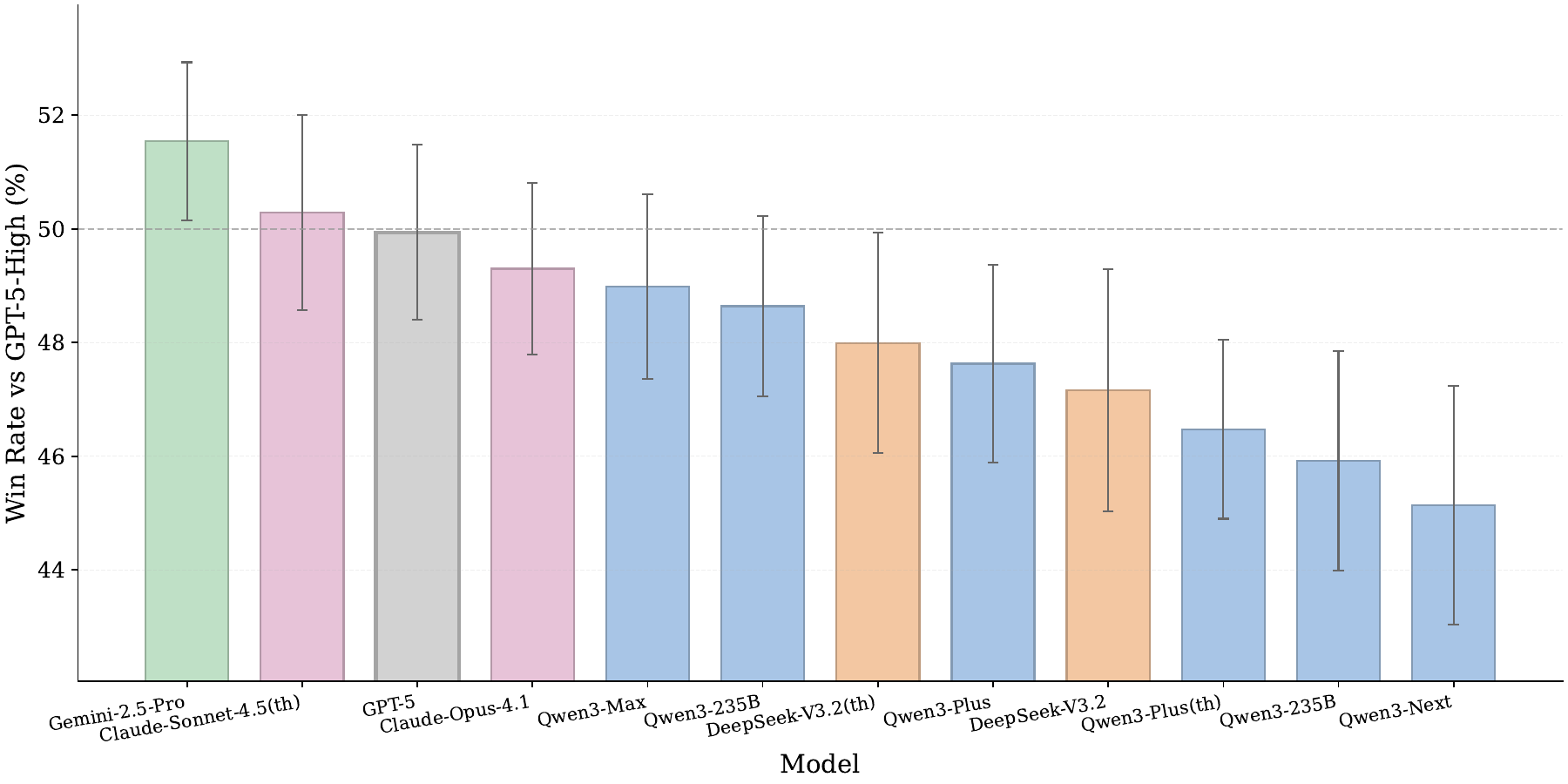}
    \caption{Arena-style open-ended evaluation results on Lao-500. Scores are win rates against GPT-5-High with 95\% bootstrap confidence intervals (CI), aggregated over two judges (Gemini-2.5-Pro and Qwen3-Max).}
    \label{fig:lao500_arena}
\end{figure}

Closed-form multiple-choice evaluation cannot fully reflect user-facing generation quality such as coherence, explanation quality, and long-form instruction following.
We therefore evaluate open-ended prompts in Lao-500 using an Arena-style pairwise protocol, which compares model outputs by preference rather than absolute scoring.
\vspace{-6pt}
\paragraph{Baseline and pairwise comparison.}
For each prompt $x_i$, we generate one response from a candidate model $M$, denoted as $y_i^{M}$, and one response from a fixed baseline model $B$ (GPT-5-High), denoted as $y_i^{B}$.
A judge model $J$ is asked to decide which response is better with respect to correctness, completeness, reasoning quality, clarity, and Lao fluency.
To mitigate potential position bias, we randomize the response ordering and repeat evaluation with swapped positions, then average the two outcomes for each prompt.
\vspace{-6pt}
\paragraph{Self-preference and judge bias control.}
To reduce potential self-preference effects, we avoid using a judge model to evaluate comparisons where it is also a candidate model whenever applicable,
and we additionally report judge-specific scores and cross-judge gaps (Appendix~\ref{sec:appendix.arena_judge_breakdown}).
This design helps mitigate correlated preferences between judge and candidate model families.
\vspace{-4pt}
\paragraph{Win-rate score.}
Let $w_i^{J}(M)\in[0,1]$ denote the (bias-corrected) win signal of model $M$ on prompt $i$ under judge $J$, where ties are assigned a fractional win. In practice we treat ties as half-wins.
We compute the judge-specific win rate of model $M$ against baseline $B$ as:
\begin{equation}
S_J(M) = \frac{1}{N}\sum_{i=1}^{N} w_i^{J}(M).
\end{equation}
This score is reported as a percentage.
\vspace{-4pt}
\paragraph{Multi-judge aggregation.}
We note that judge models may exhibit systematic preferences that correlate with model families. 
We therefore employ two judges, Gemini-2.5-Pro and Qwen3-Max, and define the overall score as the average across judges:
\begin{equation}
S(M) = \frac{1}{|\mathcal{J}|}\sum_{J\in\mathcal{J}} S_J(M),
\end{equation}
where $\mathcal{J}$ denotes the set of judges.
To mitigate single-judge bias, we aggregate scores across two independent judges and report judge-specific results in Appendix~\ref{sec:appendix.arena_judge_breakdown}.
\vspace{-4pt}
\paragraph{Bootstrap confidence intervals.}
To quantify uncertainty under a limited prompt budget, we estimate confidence intervals via bootstrap resampling over prompts.
For each bootstrap trial $t$, we sample a multiset of prompts $\mathcal{I}^{(t)}$ with replacement and recompute the score:
\vspace{-2pt}
\begin{equation}
S^{(t)}(M) = \frac{1}{|\mathcal{J}|}\sum_{J\in\mathcal{J}}
\left(\frac{1}{|\mathcal{I}^{(t)}|}\sum_{i\in \mathcal{I}^{(t)}} w_i^{J}(M)\right).
\end{equation}
The reported 95\% confidence interval is obtained from the percentile interval of the bootstrap distribution $S^{(t)}(M)$.
\vspace{-6pt}
\paragraph{Main results.}
Figure~\ref{fig:lao500_arena} shows the aggregated Lao-500 win-rate results (Avg Score) and 95\% bootstrap confidence intervals.
To make judge sensitivity explicit, we report judge-specific win rates, confidence intervals, and cross-judge gaps in Appendix~\ref{sec:appendix.arena_judge_breakdown}.
We further report judge agreement statistics and a human sanity-check subset in Appendix~\ref{sec:appendix.arena_reliability} to quantify judge sensitivity and potential bias.
\vspace{-6pt}
\paragraph{Analysis of Lao-500 results.}
We highlight three key observations.

\textbf{(1) Rankings differ from closed-form evaluation.}
Model rankings under Lao-500 are not strictly identical to Lao-7k accuracy rankings, indicating that open-ended evaluation captures additional aspects such as explanation quality and coherence beyond option selection.

\textbf{(2) Confidence intervals enable uncertainty-aware comparison.}
Models with close scores often exhibit overlapping confidence intervals, suggesting their differences may not be statistically significant under the sampled prompt set.
Non-overlapping intervals indicate robust gaps.

\textbf{(3) Judge sensitivity is non-negligible.}
While the two judges agree on major trends, we observe systematic shifts for certain model families.
For example, Qwen3-Max judge tends to assign higher win rates to Qwen-family models compared to Gemini-2.5-Pro, indicating judge-dependent preference patterns.


\section{Related Work}

\paragraph{Benchmarks for low-resource languages.}
Recent efforts have expanded evaluation resources for underrepresented languages, including African and Indic language benchmarks~\cite{adelani2021masakhanernamedentityrecognition,kakwani-etal-2020-indicnlpsuite}. 
For Southeast Asia, benchmarks have been proposed for Thai, Vietnamese, and Indonesian~\cite{yang-etal-2022-cino,xu2025cmhgdatasetbenchmarkheadline,zhang2025milic,raja2025parallelcorporamachinetranslation}. 
However, Lao remains largely absent from large-scale multidimensional evaluation suites, limiting systematic assessment of modern multilingual LLMs in this language.
\vspace{-6pt}
\paragraph{Multilingual evaluation and translation benchmarks.}
Multilingual benchmarks and large-scale MT evaluation datasets provide broad language coverage, but they often focus on translation or surface-level understanding and typically exclude Lao or provide limited Lao-specific evaluation. 
Moreover, multilingual benchmarks rarely include culturally grounded knowledge reasoning or education-aligned tasks that reflect real-world language use in Laos. 
\vspace{-6pt}
\paragraph{Dataset construction and quality assurance.}
High-quality benchmarks increasingly combine human expertise with automated verification to ensure scalability and reliability~\cite{hendrycks2021measuringmathematicalproblemsolving, luo-etal-2024-codis, wang2025mucarbenchmarkingmultilingualcrossmodal,kang2025hssbench,feng2025seeing,zheng2026should,kang2026quanteval,feng2025seeing}.
In particular, recent work highlights the importance of held-out test sets and black-box evaluation to mitigate contamination~\cite{deng-etal-2024-investigating}. 
LaoBench follows this trend by integrating expert curation with agent-assisted validation and providing a large closed-source subset.

\section{Conclusion}

We present \textbf{LaoBench}, the first large-scale multidimensional benchmark for evaluating LLMs in Lao, covering culturally grounded knowledge, K12 education, and bilingual translation.
LaoBench includes both open-source subsets for reproducible research and a large closed-source subset for secure black-box evaluation.
Experiments on Lao-7k and Lao-500 reveal that even strong multilingual models still lag behind human experts, especially on knowledge-intensive reasoning and high-fidelity translation.
We hope LaoBench will facilitate more reliable evaluation and drive progress for Lao and other underrepresented Southeast Asian languages.

\section*{Limitations}
\label{sec:limitations}
LaoBench has several limitations.
First, a large portion of the benchmark uses multiple-choice questions, which may not fully reflect open-ended reasoning ability and can allow partial gains from test-taking strategies.
Second, translation evaluation relies primarily on reference-based metrics such as BLEU, which can under-estimate valid paraphrastic translations and can be sensitive to tokenization for Lao script.
Third, Arena-style evaluation depends on LLM-based judges and a fixed baseline model, which may introduce preference bias or baseline anchoring effects, although we mitigate these via multi-judge aggregation and sanity checks. In particular, a judge model may favor outputs that resemble its own instruction style or that come from the same model family.
We partially mitigate this by using two independent judges and reporting judge-specific scores and cross-judge gaps in the appendix.
Finally, the online black-box evaluation service for Lao-10k is under active development and will be released upon publication.

\section*{Ethics Statement}
\label{sec:ethics}

\paragraph{Data sourcing and licensing.}
We collect materials from publicly accessible authoritative sources such as textbooks, curriculum guidelines, and government documents. 
For open-source releases, we only distribute derived benchmark items and do not redistribute copyrighted raw texts. 
We comply with institutional policies and ensure that released content does not contain identifiable private information.

\paragraph{Annotator welfare.}
All human annotators and reviewers are compensated at competitive local market rates. 
They receive clear task instructions, are informed of potential sensitive topics, and may opt out of sensitive content annotation.

\paragraph{Sensitive content and harm mitigation.}
Since Knowledge Application may involve political, legal, and cultural topics, we apply multi-stage sensitivity screening using both agent-assisted filtering and expert review. 
We remove or revise items that may promote hate, discrimination, misinformation, or political persuasion, and we avoid collecting or releasing personal data.

\paragraph{Intended use and reporting mechanism.}
LaoBench is intended for research and evaluation of language technologies for Lao. 
We discourage use for surveillance, targeted manipulation, or misinformation generation.



\bibliography{custom}

@inproceedings{
li2025from,
title={From Crowdsourced Data to High-quality Benchmarks: Arena-Hard and Benchbuilder Pipeline},
author={Tianle Li and Wei-Lin Chiang and Evan Frick and Lisa Dunlap and Tianhao Wu and Banghua Zhu and Joseph E. Gonzalez and Ion Stoica},
booktitle={Forty-second International Conference on Machine Learning},
year={2025},
url={https://openreview.net/forum?id=KfTf9vFvSn}
}

@misc{qwen3technicalreport,
      title={Qwen3 Technical Report}, 
      author={Qwen Team},
      year={2025},
      eprint={2505.09388},
      archivePrefix={arXiv},
      primaryClass={cs.CL},
      url={https://arxiv.org/abs/2505.09388}, 
}

@article{qwen2.5-1m,
      title={Qwen2.5-1M Technical Report}, 
      author={An Yang and Bowen Yu and Chengyuan Li and Dayiheng Liu and Fei Huang and Haoyan Huang and Jiandong Jiang and Jianhong Tu and Jianwei Zhang and Jingren Zhou and Junyang Lin and Kai Dang and Kexin Yang and Le Yu and Mei Li and Minmin Sun and Qin Zhu and Rui Men and Tao He and Weijia Xu and Wenbiao Yin and Wenyuan Yu and Xiafei Qiu and Xingzhang Ren and Xinlong Yang and Yong Li and Zhiying Xu and Zipeng Zhang},
      journal={arXiv preprint arXiv:2501.15383},
      year={2025}
}

@misc{deepseekai2024deepseekv32,
      title={DeepSeek-V3.2-Exp: Boosting Long-Context Efficiency with DeepSeek Sparse Attention}, 
      author={DeepSeek-AI},
      year={2025},
}

@misc{mistralai2024ministral8b,
  author = {{The Mistral AI Team}},
  title = {{Ministral-8B-Instruct-2410}},
  year = {2024},
  howpublished = {\url{https://huggingface.co/mistralai/Ministral-8B-Instruct-2410}},
}

@misc{Ling-mini-2.0,
  title        = {Ling-mini-2.0},
  author       = {{inclusionAI}},
  year         = {2025},
  howpublished = {\url{https://huggingface.co/inclusionAI/Ling-mini-2.0}},
}

@misc{gpt5,
  author       = {OpenAI},
  title        = {GPT-5},
  year         = {2025},
  howpublished = {\url{https://openai.com/index/introducing-gpt-5/}},
}

@misc{comanici2025gemini25pushingfrontier,
      title={Gemini 2.5: Pushing the Frontier with Advanced Reasoning, Multimodality, Long Context, and Next Generation Agentic Capabilities}, 
      author={Gheorghe Comanici and Eric Bieber and Mike Schaekermann and Ice Pasupat and Noveen Sachdeva and Inderjit Dhillon and Marcel Blistein and Ori Ram and Dan Zhang and Evan Rosen and Luke Marris and Sam Petulla and Colin Gaffney and Asaf Aharoni and Nathan Lintz and Tiago Cardal Pais and Henrik Jacobsson and Idan Szpektor and Nan-Jiang Jiang and Krishna Haridasan and Ahmed Omran and Nikunj Saunshi and Dara Bahri and Gaurav Mishra and Eric Chu and Toby Boyd and Brad Hekman and Aaron Parisi and Chaoyi Zhang and Kornraphop Kawintiranon and Tania Bedrax-Weiss and Oliver Wang and Ya Xu and Ollie Purkiss and Uri Mendlovic and Ilaï Deutel and Nam Nguyen and Adam Langley and Flip Korn and Lucia Rossazza and Alexandre Ramé and Sagar Waghmare and Helen Miller and Nathan Byrd and Ashrith Sheshan and Raia Hadsell and Sangnie Bhardwaj and Pawel Janus and Tero Rissa and Dan Horgan and Alvin Abdagic and Lior Belenki and James Allingham and Anima Singh and Theo Guidroz and Srivatsan Srinivasan and Herman Schmit and Kristen Chiafullo and Andre Elisseeff and Nilpa Jha and Prateek Kolhar and Leonard Berrada and Frank Ding and Xiance Si and Shrestha Basu Mallick and Franz Och and Sofia Erell and Eric Ni and Tejasi Latkar and Sherry Yang and Petar Sirkovic and Ziqiang Feng and Robert Leland and Rachel Hornung and Gang Wu and Charles Blundell and Hamidreza Alvari and Po-Sen Huang and Cathy Yip and Sanja Deur and Li Liu and Gabriela Surita and Pablo Duque and Dima Damen and Johnson Jia and Arthur Guez and Markus Mircea and Animesh Sinha and Alberto Magni and Paweł Stradomski and Tal Marian and Vlado Galić and Wenhu Chen and Hisham Husain and Achintya Singhal and Dominik Grewe and François-Xavier Aubet and Shuang Song and Lorenzo Blanco and Leland Rechis and Lewis Ho and Rich Munoz and Kelvin Zheng and Jessica Hamrick and Kevin Mather and Hagai Taitelbaum and Eliza Rutherford and Yun Lei and Kuangyuan Chen and Anand Shukla and Erica Moreira and Eric Doi and Berivan Isik and Nir Shabat and Dominika Rogozińska and Kashyap Kolipaka and Jason Chang and Eugen Vušak and Srinivasan Venkatachary and Shadi Noghabi and Tarun Bharti and Younghoon Jun and Aleksandr Zaks and Simon Green and Jeshwanth Challagundla and William Wong and Muqthar Mohammad and Dean Hirsch and Yong Cheng and Iftekhar Naim and Lev Proleev and Damien Vincent and Aayush Singh and Maxim Krikun and Dilip Krishnan and Zoubin Ghahramani and Aviel Atias and Rajeev Aggarwal and Christo Kirov and Dimitrios Vytiniotis and Christy Koh and Alexandra Chronopoulou and Pawan Dogra and Vlad-Doru Ion and Gladys Tyen and Jason Lee and Felix Weissenberger and Trevor Strohman and Ashwin Balakrishna and Jack Rae and Marko Velic and Raoul de Liedekerke and Oded Elyada and Wentao Yuan and Canoee Liu and Lior Shani and Sergey Kishchenko and Bea Alessio and Yandong Li and Richard Song and Sam Kwei and Orion Jankowski and Aneesh Pappu and Youhei Namiki and Yenai Ma and Nilesh Tripuraneni and Colin Cherry and Marissa Ikonomidis and Yu-Cheng Ling and Colin Ji and Beka Westberg and Auriel Wright and Da Yu and David Parkinson and Swaroop Ramaswamy and Jerome Connor and Soheil Hassas Yeganeh and Snchit Grover and George Kenwright and Lubo Litchev and Chris Apps and Alex Tomala and Felix Halim and Alex Castro-Ros and Zefei Li and Anudhyan Boral and Pauline Sho and Michal Yarom and Eric Malmi and David Klinghoffer and Rebecca Lin and Alan Ansell and Pradeep Kumar S and Shubin Zhao and Siqi Zuo and Adam Santoro and Heng-Tze Cheng and Solomon Demmessie and Yuchi Liu and Nicole Brichtova and Allie Culp and Nathaniel Braun and Dan Graur and Will Ng and Nikhil Mehta and Aaron Phillips and Patrik Sundberg and Varun Godbole and Fangyu Liu and Yash Katariya and David Rim and Mojtaba Seyedhosseini and Sean Ammirati and Jonas Valfridsson and Mahan Malihi and Timothy Knight and Andeep Toor and Thomas Lampe and Abe Ittycheriah and Lewis Chiang and Chak Yeung and Alexandre Fréchette and Jinmeng Rao and Huisheng Wang and Himanshu Srivastava and Richard Zhang and Rocky Rhodes and Ariel Brand and Dean Weesner and Ilya Figotin and Felix Gimeno and Rachana Fellinger and Pierre Marcenac and José Leal and Eyal Marcus and Victor Cotruta and Rodrigo Cabrera and Sheryl Luo and Dan Garrette and Vera Axelrod and Sorin Baltateanu and David Barker and Dongkai Chen and Horia Toma and Ben Ingram and Jason Riesa and Chinmay Kulkarni and Yujing Zhang and Hongbin Liu and Chao Wang and Martin Polacek and Will Wu and Kai Hui and Adrian N Reyes and Yi Su and Megan Barnes and Ishaan Malhi and Anfal Siddiqui and Qixuan Feng and Mihai Damaschin and Daniele Pighin and Andreas Steiner and Samuel Yang and Ramya Sree Boppana and Simeon Ivanov and Arun Kandoor and Aditya Shah and Asier Mujika and Da Huang and Christopher A. Choquette-Choo and Mohak Patel and Tianhe Yu and Toni Creswell and Jerry and Liu and Catarina Barros and Yasaman Razeghi and Aurko Roy and Phil Culliton and Binbin Xiong and Jiaqi Pan and Thomas Strohmann and Tolly Powell and Babi Seal and Doug DeCarlo and Pranav Shyam and Kaan Katircioglu and Xuezhi Wang and Cassidy Hardin and Immanuel Odisho and Josef Broder and Oscar Chang and Arun Nair and Artem Shtefan and Maura O'Brien and Manu Agarwal and Sahitya Potluri and Siddharth Goyal and Amit Jhindal and Saksham Thakur and Yury Stuken and James Lyon and Kristina Toutanova and Fangxiaoyu Feng and Austin Wu and Ben Horn and Alek Wang and Alex Cullum and Gabe Taubman and Disha Shrivastava and Chongyang Shi and Hamish Tomlinson and Roma Patel and Tao Tu and Ada Maksutaj Oflazer and Francesco Pongetti and Mingyao Yang and Adrien Ali Taïga and Vincent Perot and Nuo Wang Pierse and Feng Han and Yoel Drori and Iñaki Iturrate and Ayan Chakrabarti and Legg Yeung and Dave Dopson and Yi-ting Chen and Apoorv Kulshreshtha and Tongfei Guo and Philip Pham and Tal Schuster and Junquan Chen and Alex Polozov and Jinwei Xing and Huanjie Zhou and Praneeth Kacham and Doron Kukliansky and Antoine Miech and Sergey Yaroshenko and Ed Chi and Sholto Douglas and Hongliang Fei and Mathieu Blondel and Preethi Myla and Lior Madmoni and Xing Wu and Daniel Keysers and Kristian Kjems and Isabela Albuquerque and Lijun Yu and Joel D'sa and Michelle Plantan and Vlad Ionescu and Jaume Sanchez Elias and Abhirut Gupta and Manish Reddy Vuyyuru and Fred Alcober and Tong Zhou and Kaiyang Ji and Florian Hartmann and Subha Puttagunta and Hugo Song and Ehsan Amid and Anca Stefanoiu and Andrew Lee and Paul Pucciarelli and Emma Wang and Amit Raul and Slav Petrov and Isaac Tian and Valentin Anklin and Nana Nti and Victor Gomes and Max Schumacher and Grace Vesom and Alex Panagopoulos and Konstantinos Bousmalis and Daniel Andor and Josh Jacob and Yuan Zhang and Bill Rosgen and Matija Kecman and Matthew Tung and Alexandra Belias and Noah Goodman and Paul Covington and Brian Wieder and Nikita Saxena and Elnaz Davoodi and Muhuan Huang and Sharath Maddineni and Vincent Roulet and Folawiyo Campbell-Ajala and Pier Giuseppe Sessa and Xintian and Wu and Guangda Lai and Paul Collins and Alex Haig and Vytenis Sakenas and Xiaowei Xu and Marissa Giustina and Laurent El Shafey and Pichi Charoenpanit and Shefali Garg and Joshua Ainslie and Boone Severson and Montse Gonzalez Arenas and Shreya Pathak and Sujee Rajayogam and Jie Feng and Michiel Bakker and Sheng Li and Nevan Wichers and Jamie Rogers and Xinyang Geng and Yeqing Li and Rolf Jagerman and Chao Jia and Nadav Olmert and David Sharon and Matthew Mauger and Sandeep Mariserla and Hongxu Ma and Megha Mohabey and Kyuyeun Kim and Alek Andreev and Scott Pollom and Juliette Love and Vihan Jain and Priyanka Agrawal and Yannick Schroecker and Alisa Fortin and Manfred Warmuth and Ji Liu and Andrew Leach and Irina Blok and Ganesh Poomal Girirajan and Roee Aharoni and Benigno Uria and Andrei Sozanschi and Dan Goldberg and Lucian Ionita and Marco Tulio Ribeiro and Martin Zlocha and Vighnesh Birodkar and Sami Lachgar and Liangzhe Yuan and Himadri Choudhury and Matt Ginsberg and Fei Zheng and Gregory Dibb and Emily Graves and Swachhand Lokhande and Gabriel Rasskin and George-Cristian Muraru and Corbin Quick and Sandeep Tata and Pierre Sermanet and Aditya Chawla and Itay Karo and Yan Wang and Susan Zhang and Orgad Keller and Anca Dragan and Guolong Su and Ian Chou and Xi Liu and Yiqing Tao and Shruthi Prabhakara and Marc Wilson and Ruibo Liu and Shibo Wang and Georgie Evans and David Du and Alfonso Castaño and Gautam Prasad and Mona El Mahdy and Sebastian Gerlach and Machel Reid and Jarrod Kahn and Amir Zait and Thanumalayan Sankaranarayana Pillai and Thatcher Ulrich and Guanyu Wang and Jan Wassenberg and Efrat Farkash and Kiran Yalasangi and Congchao Wang and Maria Bauza and Simon Bucher and Ting Liu and Jun Yan and Gary Leung and Vikas Sindhwani and Parker Barnes and Avi Singh and Ivan Jurin and Jichuan Chang and Niket Kumar Bhumihar and Sivan Eiger and Gui Citovsky and Ben Withbroe and Zhang Li and Siyang Xue and Niccolò Dal Santo and Georgi Stoyanov and Yves Raimond and Steven Zheng and Yilin Gao and Vít Listík and Sławek Kwasiborski and Rachel Saputro and Adnan Ozturel and Ganesh Mallya and Kushal Majmundar and Ross West and Paul Caron and Jinliang Wei and Lluis Castrejon and Sharad Vikram and Deepak Ramachandran and Nikhil Dhawan and Jiho Park and Sara Smoot and George van den Driessche and Yochai Blau and Chase Malik and Wei Liang and Roy Hirsch and Cicero Nogueira dos Santos and Eugene Weinstein and Aäron van den Oord and Sid Lall and Nicholas FitzGerald and Zixuan Jiang and Xuan Yang and Dale Webster and Ali Elqursh and Aedan Pope and Georges Rotival and David Raposo and Wanzheng Zhu and Jeff Dean and Sami Alabed and Dustin Tran and Arushi Gupta and Zach Gleicher and Jessica Austin and Edouard Rosseel and Megh Umekar and Dipanjan Das and Yinghao Sun and Kai Chen and Karolis Misiunas and Xiang Zhou and Yixian Di and Alyssa Loo and Josh Newlan and Bo Li and Vinay Ramasesh and Ying Xu and Alex Chen and Sudeep Gandhe and Radu Soricut and Nikita Gupta and Shuguang Hu and Seliem El-Sayed and Xavier Garcia and Idan Brusilovsky and Pu-Chin Chen and Andrew Bolt and Lu Huang and Alex Gurney and Zhiying Zhang and Alexander Pritzel and Jarek Wilkiewicz and Bryan Seybold and Bhargav Kanagal Shamanna and Felix Fischer and Josef Dean and Karan Gill and Ross Mcilroy and Abhishek Bhowmick and Jeremy Selier and Antoine Yang and Derek Cheng and Vladimir Magay and Jie Tan and Dhriti Varma and Christian Walder and Tomas Kocisky and Ryo Nakashima and Paul Natsev and Mike Kwong and Ionel Gog and Chiyuan Zhang and Sander Dieleman and Thomas Jimma and Andrey Ryabtsev and Siddhartha Brahma and David Steiner and Dayou Du and Ante Žužul and Mislav Žanić and Mukund Raghavachari and Willi Gierke and Zeyu Zheng and Dessie Petrova and Yann Dauphin and Yuchuan Liu and Ido Kessler and Steven Hand and Chris Duvarney and Seokhwan Kim and Hyo Lee and Léonard Hussenot and Jeffrey Hui and Josh Smith and Deepali Jain and Jiawei Xia and Gaurav Singh Tomar and Keyvan Amiri and Du Phan and Fabian Fuchs and Tobias Weyand and Nenad Tomasev and Alexandra Cordell and Xin Liu and Jonathan Mallinson and Pankaj Joshi and Andy Crawford and Arun Suggala and Steve Chien and Nick Fernando and Mariella Sanchez-Vargas and Duncan Williams and Phil Crone and Xiyang Luo and Igor Karpov and Jyn Shan and Terry Thurk and Robin Strudel and Paul Voigtlaender and Piyush Patil and Tim Dozat and Ali Khodaei and Sahil Singla and Piotr Ambroszczyk and Qiyin Wu and Yifan Chang and Brian Roark and Chaitra Hegde and Tianli Ding and Angelos Filos and Zhongru Wu and André Susano Pinto and Shuang Liu and Saarthak Khanna and Aditya Pandey and Siobhan Mcloughlin and Qiujia Li and Sam Haves and Allan Zhou and Elena Buchatskaya and Isabel Leal and Peter de Boursac and Nami Akazawa and Nina Anderson and Terry Chen and Krishna Somandepalli and Chen Liang and Sheela Goenka and Stephanie Winkler and Alexander Grushetsky and Yifan Ding and Jamie Smith and Fan Ye and Jordi Pont-Tuset and Eric Li and Ruichao Li and Tomer Golany and Dawid Wegner and Tao Jiang and Omer Barak and Yuan Shangguan and Eszter Vértes and Renee Wong and Jörg Bornschein and Alex Tudor and Michele Bevilacqua and Tom Schaul and Ankit Singh Rawat and Yang Zhao and Kyriakos Axiotis and Lei Meng and Cory McLean and Jonathan Lai and Jennifer Beattie and Nate Kushman and Yaxin Liu and Blair Kutzman and Fiona Lang and Jingchen Ye and Praneeth Netrapalli and Pushkar Mishra and Myriam Khan and Megha Goel and Rob Willoughby and David Tian and Honglei Zhuang and JD Chen and Zak Tsai and Tasos Kementsietsidis and Arjun Khare and James Keeling and Keyang Xu and Nathan Waters and Florent Altché and Ashok Popat and Bhavishya Mittal and David Saxton and Dalia El Badawy and Michael Mathieu and Zheng Zheng and Hao Zhou and Nishant Ranka and Richard Shin and Qingnan Duan and Tim Salimans and Ioana Mihailescu and Uri Shaham and Ming-Wei Chang and Yannis Assael and Nishanth Dikkala and Martin Izzard and Vincent Cohen-Addad and Cat Graves and Vlad Feinberg and Grace Chung and DJ Strouse and Danny Karmon and Sahand Sharifzadeh and Zoe Ashwood and Khiem Pham and Jon Blanton and Alex Vasiloff and Jarred Barber and Mark Geller and Aurick Zhou and Fedir Zubach and Tzu-Kuo Huang and Lei Zhang and Himanshu Gupta and Matt Young and Julia Proskurnia and Ronny Votel and Valentin Gabeur and Gabriel Barcik and Aditya Tripathi and Hongkun Yu and Geng Yan and Beer Changpinyo and Filip Pavetić and Amy Coyle and Yasuhisa Fujii and Jorge Gonzalez Mendez and Tianhao Zhou and Harish Rajamani and Blake Hechtman and Eddie Cao and Da-Cheng Juan and Yi-Xuan Tan and Valentin Dalibard and Yilun Du and Natalie Clay and Kaisheng Yao and Wenhao Jia and Dimple Vijaykumar and Yuxiang Zhou and Xinyi Bai and Wei-Chih Hung and Steven Pecht and Georgi Todorov and Nikhil Khadke and Pramod Gupta and Preethi Lahoti and Arnaud Autef and Karthik Duddu and James Lee-Thorp and Alexander Bykovsky and Tautvydas Misiunas and Sebastian Flennerhag and Santhosh Thangaraj and Jed McGiffin and Zack Nado and Markus Kunesch and Andreas Noever and Amir Hertz and Marco Liang and Victor Stone and Evan Palmer and Samira Daruki and Arijit Pramanik and Siim Põder and Austin Kyker and Mina Khan and Evgeny Sluzhaev and Marvin Ritter and Avraham Ruderman and Wenlei Zhou and Chirag Nagpal and Kiran Vodrahalli and George Necula and Paul Barham and Ellie Pavlick and Jay Hartford and Izhak Shafran and Long Zhao and Maciej Mikuła and Tom Eccles and Hidetoshi Shimokawa and Kanav Garg and Luke Vilnis and Hanwen Chen and Ilia Shumailov and Kuang-Huei Lee and Abdelrahman Abdelhamed and Meiyan Xie and Vered Cohen and Ester Hlavnova and Dan Malkin and Chawin Sitawarin and James Lottes and Pauline Coquinot and Tianli Yu and Sandeep Kumar and Jingwei Zhang and Aroma Mahendru and Zafarali Ahmed and James Martens and Tao Chen and Aviel Boag and Daiyi Peng and Coline Devin and Arseniy Klimovskiy and Mary Phuong and Danny Vainstein and Jin Xie and Bhuvana Ramabhadran and Nathan Howard and Xinxin Yu and Gitartha Goswami and Jingyu Cui and Sam Shleifer and Mario Pinto and Chih-Kuan Yeh and Ming-Hsuan Yang and Sara Javanmardi and Dan Ethier and Chace Lee and Jordi Orbay and Suyog Kotecha and Carla Bromberg and Pete Shaw and James Thornton and Adi Gerzi Rosenthal and Shane Gu and Matt Thomas and Ian Gemp and Aditya Ayyar and Asahi Ushio and Aarush Selvan and Joel Wee and Chenxi Liu and Maryam Majzoubi and Weiren Yu and Jake Abernethy and Tyler Liechty and Renke Pan and Hoang Nguyen and Qiong and Hu and Sarah Perrin and Abhinav Arora and Emily Pitler and Weiyi Wang and Kaushik Shivakumar and Flavien Prost and Ben Limonchik and Jing Wang and Yi Gao and Timothee Cour and Shyamal Buch and Huan Gui and Maria Ivanova and Philipp Neubeck and Kelvin Chan and Lucy Kim and Huizhong Chen and Naman Goyal and Da-Woon Chung and Lu Liu and Yao Su and Anastasia Petrushkina and Jiajun Shen and Armand Joulin and Yuanzhong Xu and Stein Xudong Lin and Yana Kulizhskaya and Ciprian Chelba and Shobha Vasudevan and Eli Collins and Vasilisa Bashlovkina and Tony Lu and Doug Fritz and Jongbin Park and Yanqi Zhou and Chen Su and Richard Tanburn and Mikhail Sushkov and Mitchelle Rasquinha and Jinning Li and Jennifer Prendki and Yiming Li and Pallavi LV and Shriya Sharma and Hen Fitoussi and Hui Huang and Andrew Dai and Phuong Dao and Mike Burrows and Henry Prior and Danfeng Qin and Golan Pundak and Lars Lowe Sjoesund and Art Khurshudov and Zhenkai Zhu and Albert Webson and Elizabeth Kemp and Tat Tan and Saurabh Agrawal and Susie Sargsyan and Liqun Cheng and Jim Stephan and Tom Kwiatkowski and David Reid and Arunkumar Byravan and Assaf Hurwitz Michaely and Nicolas Heess and Luowei Zhou and Sonam Goenka and Viral Carpenter and Anselm Levskaya and Bo Wang and Reed Roberts and Rémi Leblond and Sharat Chikkerur and Stav Ginzburg and Max Chang and Robert Riachi and Chuqiao and Xu and Zalán Borsos and Michael Pliskin and Julia Pawar and Morgane Lustman and Hannah Kirkwood and Ankit Anand and Aditi Chaudhary and Norbert Kalb and Kieran Milan and Sean Augenstein and Anna Goldie and Laurel Prince and Karthik Raman and Yanhua Sun and Vivian Xia and Aaron Cohen and Zhouyuan Huo and Josh Camp and Seher Ellis and Lukas Zilka and David Vilar Torres and Lisa Patel and Sho Arora and Betty Chan and Jonas Adler and Kareem Ayoub and Jacky Liang and Fayaz Jamil and Jiepu Jiang and Simon Baumgartner and Haitian Sun and Yael Karov and Yaroslav Akulov and Hui Zheng and Irene Cai and Claudio Fantacci and James Rubin and Alex Rav Acha and Mengchao Wang and Nina D'Souza and Rohit Sathyanarayana and Shengyang Dai and Simon Rowe and Andrey Simanovsky and Omer Goldman and Yuheng Kuang and Xiaoyue Pan and Andrew Rosenberg and Tania Rojas-Esponda and Praneet Dutta and Amy Zeng and Irina Jurenka and Greg Farquhar and Yamini Bansal and Shariq Iqbal and Becca Roelofs and Ga-Young Joung and Parker Beak and Changwan Ryu and Ryan Poplin and Yan Wu and Jean-Baptiste Alayrac and Senaka Buthpitiya and Olaf Ronneberger and Caleb Habtegebriel and Wei Li and Paul Cavallaro and Aurora Wei and Guy Bensky and Timo Denk and Harish Ganapathy and Jeff Stanway and Pratik Joshi and Francesco Bertolini and Jessica Lo and Olivia Ma and Zachary Charles and Geta Sampemane and Himanshu Sahni and Xu Chen and Harry Askham and David Gaddy and Peter Young and Jiewen Tan and Matan Eyal and Arthur Bražinskas and Li Zhong and Zhichun Wu and Mark Epstein and Kai Bailey and Andrew Hard and Kamyu Lee and Sasha Goldshtein and Alex Ruiz and Mohammed Badawi and Matthias Lochbrunner and JK Kearns and Ashley Brown and Fabio Pardo and Theophane Weber and Haichuan Yang and Pan-Pan Jiang and Berkin Akin and Zhao Fu and Marcus Wainwright and Chi Zou and Meenu Gaba and Pierre-Antoine Manzagol and Wendy Kan and Yang Song and Karina Zainullina and Rui Lin and Jeongwoo Ko and Salil Deshmukh and Apoorv Jindal and James Svensson and Divya Tyam and Heri Zhao and Christine Kaeser-Chen and Scott Baird and Pooya Moradi and Jamie Hall and Qiuchen Guo and Vincent Tsang and Bowen Liang and Fernando Pereira and Suhas Ganesh and Ivan Korotkov and Jakub Adamek and Sridhar Thiagarajan and Vinh Tran and Charles Chen and Chris Tar and Sanil Jain and Ishita Dasgupta and Taylan Bilal and David Reitter and Kai Zhao and Giulia Vezzani and Yasmin Gehman and Pulkit Mehta and Lauren Beltrone and Xerxes Dotiwalla and Sergio Guadarrama and Zaheer Abbas and Stefani Karp and Petko Georgiev and Chun-Sung Ferng and Marc Brockschmidt and Liqian Peng and Christoph Hirnschall and Vikas Verma and Yingying Bi and Ying Xiao and Avigail Dabush and Kelvin Xu and Phil Wallis and Randall Parker and Qifei Wang and Yang Xu and Ilkin Safarli and Dinesh Tewari and Yin Zhang and Seungyeon Kim and Andrea Gesmundo and Mackenzie Thomas and Sergey Levi and Ahmed Chowdhury and Kanishka Rao and Peter Garst and Sam Conway-Rahman and Helen Ran and Kay McKinney and Zhisheng Xiao and Wenhao Yu and Rohan Agrawal and Axel Stjerngren and Catalin Ionescu and Jingjing Chen and Vivek Sharma and Justin Chiu and Fei Liu and Ken Franko and Clayton Sanford and Xingyu Cai and Paul Michel and Sanjay Ganapathy and Jane Labanowski and Zachary Garrett and Ben Vargas and Sean Sun and Bryan Gale and Thomas Buschmann and Guillaume Desjardins and Nimesh Ghelani and Palak Jain and Mudit Verma and Chulayuth Asawaroengchai and Julian Eisenschlos and Jitendra Harlalka and Hideto Kazawa and Don Metzler and Joshua Howland and Ying Jian and Jake Ades and Viral Shah and Tynan Gangwani and Seungji Lee and Roman Ring and Steven M. Hernandez and Dean Reich and Amer Sinha and Ashutosh Sathe and Joe Kovac and Ashleah Gill and Ajay Kannan and Andrea D'olimpio and Martin Sevenich and Jay Whang and Been Kim and Khe Chai Sim and Jilin Chen and Jiageng Zhang and Shuba Lall and Yossi Matias and Bill Jia and Abe Friesen and Sara Nasso and Ashish Thapliyal and Bryan Perozzi and Ting Yu and Anna Shekhawat and Safeen Huda and Peter Grabowski and Eric Wang and Ashwin Sreevatsa and Hilal Dib and Mehadi Hassen and Parker Schuh and Vedrana Milutinovic and Chris Welty and Michael Quinn and Ali Shah and Bangju Wang and Gabe Barth-Maron and Justin Frye and Natalie Axelsson and Tao Zhu and Yukun Ma and Irene Giannoumis and Hanie Sedghi and Chang Ye and Yi Luan and Kevin Aydin and Bilva Chandra and Vivek Sampathkumar and Ronny Huang and Victor Lavrenko and Ahmed Eleryan and Zhi Hong and Steven Hansen and Sara Mc Carthy and Bidisha Samanta and Domagoj Ćevid and Xin Wang and Fangtao Li and Michael Voznesensky and Matt Hoffman and Andreas Terzis and Vikash Sehwag and Gil Fidel and Luheng He and Mu Cai and Yanzhang He and Alex Feng and Martin Nikoltchev and Samrat Phatale and Jason Chase and Rory Lawton and Ming Zhang and Tom Ouyang and Manuel Tragut and Mehdi Hafezi Manshadi and Arjun Narayanan and Jiaming Shen and Xu Gao and Tolga Bolukbasi and Nick Roy and Xin Li and Daniel Golovin and Liviu Panait and Zhen Qin and Guangxing Han and Thomas Anthony and Sneha Kudugunta and Viorica Patraucean and Aniket Ray and Xinyun Chen and Xiaochen Yang and Tanuj Bhatia and Pranav Talluri and Alex Morris and Andrija Ražnatović and Bethanie Brownfield and James An and Sheng Peng and Patrick Kane and Ce Zheng and Nico Duduta and Joshua Kessinger and James Noraky and Siqi Liu and Keran Rong and Petar Veličković and Keith Rush and Alex Goldin and Fanny Wei and Shiva Mohan Reddy Garlapati and Caroline Pantofaru and Okwan Kwon and Jianmo Ni and Eric Noland and Julia Di Trapani and Françoise Beaufays and Abhijit Guha Roy and Yinlam Chow and Aybuke Turker and Geoffrey Cideron and Lantao Mei and Jon Clark and Qingyun Dou and Matko Bošnjak and Ralph Leith and Yuqing Du and Amir Yazdanbakhsh and Milad Nasr and Chester Kwak and Suraj Satishkumar Sheth and Alex Kaskasoli and Ankesh Anand and Balaji Lakshminarayanan and Sammy Jerome and David Bieber and Chun-Te Chu and Alexandre Senges and Tianxiao Shen and Mukund Sridhar and Ndaba Ndebele and Benjamin Beyret and Shakir Mohamed and Mia Chen and Markus Freitag and Jiaxian Guo and Luyang Liu and Paul Roit and Heng Chen and Shen Yan and Tom Stone and JD Co-Reyes and Jeremy Cole and Salvatore Scellato and Shekoofeh Azizi and Hadi Hashemi and Alicia Jin and Anand Iyer and Marcella Valentine and András György and Arun Ahuja and Daniel Hernandez Diaz and Chen-Yu Lee and Nathan Clement and Weize Kong and Drew Garmon and Ishaan Watts and Kush Bhatia and Khyatti Gupta and Matt Miecnikowski and Hugo Vallet and Ankur Taly and Edward Loper and Saket Joshi and James Atwood and Jo Chick and Mark Collier and Fotis Iliopoulos and Ryan Trostle and Beliz Gunel and Ramiro Leal-Cavazos and Arnar Mar Hrafnkelsson and Michael Guzman and Xiaoen Ju and Andy Forbes and Jesse Emond and Kushal Chauhan and Ben Caine and Li Xiao and Wenjun Zeng and Alexandre Moufarek and Daniel Murphy and Maya Meng and Nitish Gupta and Felix Riedel and Anil Das and Elijah Lawal and Shashi Narayan and Tiberiu Sosea and James Swirhun and Linda Friso and Behnam Neyshabur and Jing Lu and Sertan Girgin and Michael Wunder and Edouard Yvinec and Aroonalok Pyne and Victor Carbune and Shruti Rijhwani and Yang Guo and Tulsee Doshi and Anton Briukhov and Max Bain and Ayal Hitron and Xuanhui Wang and Ashish Gupta and Ke Chen and Cosmo Du and Weiyang Zhang and Dhruv Shah and Arjun Akula and Max Dylla and Ashyana Kachra and Weicheng Kuo and Tingting Zou and Lily Wang and Luyao Xu and Jifan Zhu and Justin Snyder and Sachit Menon and Orhan Firat and Igor Mordatch and Yuan Yuan and Natalia Ponomareva and Rory Blevins and Lawrence Moore and Weijun Wang and Phil Chen and Martin Scholz and Artur Dwornik and Jason Lin and Sicheng Li and Diego Antognini and Te I and Xiaodan Song and Matt Miller and Uday Kalra and Adam Raveret and Oscar Akerlund and Felix Wu and Andrew Nystrom and Namrata Godbole and Tianqi Liu and Hannah DeBalsi and Jewel Zhao and Buhuang Liu and Avi Caciularu and Lauren Lax and Urvashi Khandelwal and Victoria Langston and Eric Bailey and Silvio Lattanzi and Yufei Wang and Neel Kovelamudi and Sneha Mondal and Guru Guruganesh and Nan Hua and Ofir Roval and Paweł Wesołowski and Rishikesh Ingale and Jonathan Halcrow and Tim Sohn and Christof Angermueller and Bahram Raad and Eli Stickgold and Eva Lu and Alec Kosik and Jing Xie and Timothy Lillicrap and Austin Huang and Lydia Lihui Zhang and Dominik Paulus and Clement Farabet and Alex Wertheim and Bing Wang and Rishabh Joshi and Chu-ling Ko and Yonghui Wu and Shubham Agrawal and Lily Lin and XiangHai Sheng and Peter Sung and Tyler Breland-King and Christina Butterfield and Swapnil Gawde and Sumeet Singh and Qiao Zhang and Raj Apte and Shilpa Shetty and Adrian Hutter and Tao Li and Elizabeth Salesky and Federico Lebron and Jonni Kanerva and Michela Paganini and Arthur Nguyen and Rohith Vallu and Jan-Thorsten Peter and Sarmishta Velury and David Kao and Jay Hoover and Anna Bortsova and Colton Bishop and Shoshana Jakobovits and Alessandro Agostini and Alekh Agarwal and Chang Liu and Charles Kwong and Sasan Tavakkol and Ioana Bica and Alex Greve and Anirudh GP and Jake Marcus and Le Hou and Tom Duerig and Rivka Moroshko and Dave Lacey and Andy Davis and Julien Amelot and Guohui Wang and Frank Kim and Theofilos Strinopoulos and Hui Wan and Charline Le Lan and Shankar Krishnan and Haotian Tang and Peter Humphreys and Junwen Bai and Idan Heimlich Shtacher and Diego Machado and Chenxi Pang and Ken Burke and Dangyi Liu and Renga Aravamudhan and Yue Song and Ed Hirst and Abhimanyu Singh and Brendan Jou and Liang Bai and Francesco Piccinno and Chuyuan Kelly Fu and Robin Alazard and Barak Meiri and Daniel Winter and Charlie Chen and Mingda Zhang and Jens Heitkaemper and John Lambert and Jinhyuk Lee and Alexander Frömmgen and Sergey Rogulenko and Pranav Nair and Paul Niemczyk and Anton Bulyenov and Bibo Xu and Hadar Shemtov and Morteza Zadimoghaddam and Serge Toropov and Mateo Wirth and Hanjun Dai and Sreenivas Gollapudi and Daniel Zheng and Alex Kurakin and Chansoo Lee and Kalesha Bullard and Nicolas Serrano and Ivana Balazevic and Yang Li and Johan Schalkwyk and Mark Murphy and Mingyang Zhang and Kevin Sequeira and Romina Datta and Nishant Agrawal and Charles Sutton and Nithya Attaluri and Mencher Chiang and Wael Farhan and Gregory Thornton and Kate Lin and Travis Choma and Hung Nguyen and Kingshuk Dasgupta and Dirk Robinson and Iulia Comşa and Michael Riley and Arjun Pillai and Basil Mustafa and Ben Golan and Amir Zandieh and Jean-Baptiste Lespiau and Billy Porter and David Ross and Sujeevan Rajayogam and Mohit Agarwal and Subhashini Venugopalan and Bobak Shahriari and Qiqi Yan and Hao Xu and Taylor Tobin and Pavel Dubov and Hongzhi Shi and Adrià Recasens and Anton Kovsharov and Sebastian Borgeaud and Lucio Dery and Shanthal Vasanth and Elena Gribovskaya and Linhai Qiu and Mahdis Mahdieh and Wojtek Skut and Elizabeth Nielsen and CJ Zheng and Adams Yu and Carrie Grimes Bostock and Shaleen Gupta and Aaron Archer and Chris Rawles and Elinor Davies and Alexey Svyatkovskiy and Tomy Tsai and Yoni Halpern and Christian Reisswig and Bartek Wydrowski and Bo Chang and Joan Puigcerver and Mor Hazan Taege and Jian Li and Eva Schnider and Xinjian Li and Dragos Dena and Yunhan Xu and Umesh Telang and Tianze Shi and Heiga Zen and Kyle Kastner and Yeongil Ko and Neesha Subramaniam and Aviral Kumar and Pete Blois and Zhuyun Dai and John Wieting and Yifeng Lu and Yoel Zeldes and Tian Xie and Anja Hauth and Alexandru Ţifrea and Yuqi Li and Sam El-Husseini and Dan Abolafia and Howard Zhou and Wen Ding and Sahra Ghalebikesabi and Carlos Guía and Andrii Maksai and Ágoston Weisz and Sercan Arik and Nick Sukhanov and Aga Świetlik and Xuhui Jia and Luo Yu and Weiyue Wang and Mark Brand and Dawn Bloxwich and Sean Kirmani and Zhe Chen and Alec Go and Pablo Sprechmann and Nithish Kannen and Alen Carin and Paramjit Sandhu and Isabel Edkins and Leslie Nooteboom and Jai Gupta and Loren Maggiore and Javad Azizi and Yael Pritch and Pengcheng Yin and Mansi Gupta and Danny Tarlow and Duncan Smith and Desi Ivanov and Mohammad Babaeizadeh and Ankita Goel and Satish Kambala and Grace Chu and Matej Kastelic and Michelle Liu and Hagen Soltau and Austin Stone and Shivani Agrawal and Min Kim and Kedar Soparkar and Srinivas Tadepalli and Oskar Bunyan and Rachel Soh and Arvind Kannan and DY Kim and Blake JianHang Chen and Afief Halumi and Sudeshna Roy and Yulong Wang and Olcan Sercinoglu and Gena Gibson and Sijal Bhatnagar and Motoki Sano and Daniel von Dincklage and Qingchun Ren and Blagoj Mitrevski and Mirek Olšák and Jennifer She and Carl Doersch and Jilei and Wang and Bingyuan Liu and Qijun Tan and Tamar Yakar and Tris Warkentin and Alex Ramirez and Carl Lebsack and Josh Dillon and Rajiv Mathews and Tom Cobley and Zelin Wu and Zhuoyuan Chen and Jon Simon and Swaroop Nath and Tara Sainath and Alexei Bendebury and Ryan Julian and Bharath Mankalale and Daria Ćurko and Paulo Zacchello and Adam R. Brown and Kiranbir Sodhia and Heidi Howard and Sergi Caelles and Abhinav Gupta and Gareth Evans and Anna Bulanova and Lesley Katzen and Roman Goldenberg and Anton Tsitsulin and Joe Stanton and Benoit Schillings and Vitaly Kovalev and Corey Fry and Rushin Shah and Kuo Lin and Shyam Upadhyay and Cheng Li and Soroush Radpour and Marcello Maggioni and Jing Xiong and Lukas Haas and Jenny Brennan and Aishwarya Kamath and Nikolay Savinov and Arsha Nagrani and Trevor Yacovone and Ryan Kappedal and Kostas Andriopoulos and Li Lao and YaGuang Li and Grigory Rozhdestvenskiy and Kazuma Hashimoto and Andrew Audibert and Sophia Austin and Daniel Rodriguez and Anian Ruoss and Garrett Honke and Deep Karkhanis and Xi Xiong and Qing Wei and James Huang and Zhaoqi Leng and Vittal Premachandran and Stan Bileschi and Georgios Evangelopoulos and Thomas Mensink and Jay Pavagadhi and Denis Teplyashin and Paul Chang and Linting Xue and Garrett Tanzer and Sally Goldman and Kaushal Patel and Shixin Li and Jeremy Wiesner and Ivy Zheng and Ian Stewart-Binks and Jie Han and Zhi Li and Liangchen Luo and Karel Lenc and Mario Lučić and Fuzhao Xue and Ryan Mullins and Alexey Guseynov and Chung-Ching Chang and Isaac Galatzer-Levy and Adam Zhang and Garrett Bingham and Grace Hu and Ale Hartman and Yue Ma and Jordan Griffith and Alex Irpan and Carey Radebaugh and Summer Yue and Lijie Fan and Victor Ungureanu and Christina Sorokin and Hannah Teufel and Peiran Li and Rohan Anil and Dimitris Paparas and Todd Wang and Chu-Cheng Lin and Hui Peng and Megan Shum and Goran Petrovic and Demetra Brady and Richard Nguyen and Klaus Macherey and Zhihao Li and Harman Singh and Madhavi Yenugula and Mariko Iinuma and Xinyi Chen and Kavya Kopparapu and Alexey Stern and Shachi Dave and Chandu Thekkath and Florence Perot and Anurag Kumar and Fangda Li and Yang Xiao and Matthew Bilotti and Mohammad Hossein Bateni and Isaac Noble and Lisa Lee and Amelio Vázquez-Reina and Julian Salazar and Xiaomeng Yang and Boyu Wang and Ela Gruzewska and Anand Rao and Sindhu Raghuram and Zheng Xu and Eyal Ben-David and Jieru Mei and Sid Dalmia and Zhaoyi Zhang and Yuchen Liu and Gagan Bansal and Helena Pankov and Steven Schwarcz and Andrea Burns and Christine Chan and Sumit Sanghai and Ricky Liang and Ethan Liang and Antoine He and Amy Stuart and Arun Narayanan and Yukun Zhu and Christian Frank and Bahar Fatemi and Amit Sabne and Oran Lang and Indro Bhattacharya and Shane Settle and Maria Wang and Brendan McMahan and Andrea Tacchetti and Livio Baldini Soares and Majid Hadian and Serkan Cabi and Timothy Chung and Nikita Putikhin and Gang Li and Jeremy Chen and Austin Tarango and Henryk Michalewski and Mehran Kazemi and Hussain Masoom and Hila Sheftel and Rakesh Shivanna and Archita Vadali and Ramona Comanescu and Doug Reid and Joss Moore and Arvind Neelakantan and Michaël Sander and Jonathan Herzig and Aviv Rosenberg and Mostafa Dehghani and JD Choi and Michael Fink and Reid Hayes and Eric Ge and Shitao Weng and Chia-Hua Ho and John Karro and Kalpesh Krishna and Lam Nguyen Thiet and Amy Skerry-Ryan and Daniel Eppens and Marco Andreetto and Navin Sarma and Silvano Bonacina and Burcu Karagol Ayan and Megha Nawhal and Zhihao Shan and Mike Dusenberry and Shantanu Thakoor and Sagar Gubbi and Duc Dung Nguyen and Reut Tsarfaty and Samuel Albanie and Jovana Mitrović and Meet Gandhi and Bo-Juen Chen and Alessandro Epasto and Georgi Stephanov and Ye Jin and Samuel Gehman and Aida Amini and Jack Weber and Feryal Behbahani and Shawn Xu and Miltos Allamanis and Xi Chen and Myle Ott and Claire Sha and Michal Jastrzebski and Hang Qi and David Greene and Xinyi Wu and Abodunrinwa Toki and Daniel Vlasic and Jane Shapiro and Ragha Kotikalapudi and Zhe Shen and Takaaki Saeki and Sirui Xie and Albin Cassirer and Shikhar Bharadwaj and Tatsuya Kiyono and Srinadh Bhojanapalli and Elan Rosenfeld and Sam Ritter and Jieming Mao and João Gabriel Oliveira and Zoltan Egyed and Bernd Bandemer and Emilio Parisotto and Keisuke Kinoshita and Juliette Pluto and Petros Maniatis and Steve Li and Yaohui Guo and Golnaz Ghiasi and Jean Tarbouriech and Srimon Chatterjee and Julie Jin and Katrina and Xu and Jennimaria Palomaki and Séb Arnold and Madhavi Sewak and Federico Piccinini and Mohit Sharma and Ben Albrecht and Sean Purser-haskell and Ashwin Vaswani and Chongyan Chen and Matheus Wisniewski and Qin Cao and John Aslanides and Nguyet Minh Phu and Maximilian Sieb and Lauren Agubuzu and Anne Zheng and Daniel Sohn and Marco Selvi and Anders Andreassen and Krishan Subudhi and Prem Eruvbetine and Oliver Woodman and Tomas Mery and Sebastian Krause and Xiaoqi Ren and Xiao Ma and Jincheng Luo and Dawn Chen and Wei Fan and Henry Griffiths and Christian Schuler and Alice Li and Shujian Zhang and Jean-Michel Sarr and Shixin Luo and Riccardo Patana and Matthew Watson and Dani Naboulsi and Michael Collins and Sailesh Sidhwani and Emiel Hoogeboom and Sharon Silver and Emily Caveness and Xiaokai Zhao and Mikel Rodriguez and Maxine Deines and Libin Bai and Patrick Griffin and Marco Tagliasacchi and Emily Xue and Spandana Raj Babbula and Bo Pang and Nan Ding and Gloria Shen and Elijah Peake and Remi Crocker and Shubha Srinivas Raghvendra and Danny Swisher and Woohyun Han and Richa Singh and Ling Wu and Vladimir Pchelin and Tsendsuren Munkhdalai and Dana Alon and Geoff Bacon and Efren Robles and Jannis Bulian and Melvin Johnson and George Powell and Felipe Tiengo Ferreira and Yaoyiran Li and Frederik Benzing and Mihajlo Velimirović and Hubert Soyer and William Kong and Tony and Nguyên and Zhen Yang and Jeremiah Liu and Joost van Amersfoort and Daniel Gillick and Baochen Sun and Nathalie Rauschmayr and Katie Zhang and Serena Zhan and Tao Zhou and Alexey Frolov and Chengrun Yang and Denis Vnukov and Louis Rouillard and Hongji Li and Amol Mandhane and Nova Fallen and Rajesh Venkataraman and Clara Huiyi Hu and Jennifer Brennan and Jenny Lee and Jerry Chang and Martin Sundermeyer and Zhufeng Pan and Rosemary Ke and Simon Tong and Alex Fabrikant and William Bono and Jindong Gu and Ryan Foley and Yiran Mao and Manolis Delakis and Dhruva Bhaswar and Roy Frostig and Nick Li and Avital Zipori and Cath Hope and Olga Kozlova and Swaroop Mishra and Josip Djolonga and Craig Schiff and Majd Al Merey and Eleftheria Briakou and Peter Morgan and Andy Wan and Avinatan Hassidim and RJ Skerry-Ryan and Kuntal Sengupta and Mary Jasarevic and Praveen Kallakuri and Paige Kunkle and Hannah Brennan and Tom Lieber and Hassan Mansoor and Julian Walker and Bing Zhang and Annie Xie and Goran Žužić and Adaeze Chukwuka and Alex Druinsky and Donghyun Cho and Rui Yao and Ferjad Naeem and Shiraz Butt and Eunyoung Kim and Zhipeng Jia and Mandy Jordan and Adam Lelkes and Mark Kurzeja and Sophie Wang and James Zhao and Andrew Over and Abhishek Chakladar and Marcel Prasetya and Neha Jha and Sriram Ganapathy and Yale Cong and Prakash Shroff and Carl Saroufim and Sobhan Miryoosefi and Mohamed Hammad and Tajwar Nasir and Weijuan Xi and Yang Gao and Young Maeng and Ben Hora and Chin-Yi Cheng and Parisa Haghani and Yoad Lewenberg and Caden Lu and Martin Matysiak and Naina Raisinghani and Huiyu Wang and Lexi Baugher and Rahul Sukthankar and Minh Giang and John Schultz and Noah Fiedel and Minmin Chen and Cheng-Chun Lee and Tapomay Dey and Hao Zheng and Shachi Paul and Celine Smith and Andy Ly and Yicheng Wang and Rishabh Bansal and Bartek Perz and Susanna Ricco and Stasha Blank and Vaishakh Keshava and Deepak Sharma and Marvin Chow and Kunal Lad and Komal Jalan and Simon Osindero and Craig Swanson and Jacob Scott and Anastasija Ilić and Xiaowei Li and Siddhartha Reddy Jonnalagadda and Afzal Shama Soudagar and Yan Xiong and Bat-Orgil Batsaikhan and Daniel Jarrett and Naveen Kumar and Maulik Shah and Matt Lawlor and Austin Waters and Mark Graham and Rhys May and Sabela Ramos and Sandra Lefdal and Zeynep Cankara and Nacho Cano and Brendan O'Donoghue and Jed Borovik and Frederick Liu and Jordan Grimstad and Mahmoud Alnahlawi and Katerina Tsihlas and Tom Hudson and Nikolai Grigorev and Yiling Jia and Terry Huang and Tobenna Peter Igwe and Sergei Lebedev and Xiaodan Tang and Igor Krivokon and Frankie Garcia and Melissa Tan and Eric Jia and Peter Stys and Shikhar Vashishth and Yu Liang and Balaji Venkatraman and Chenjie Gu and Anastasios Kementsietsidis and Chen Zhu and Junehyuk Jung and Yunfei Bai and Mohammad Javad Hosseini and Faruk Ahmed and Aditya Gupta and Xin Yuan and Shereen Ashraf and Shitij Nigam and Gautam Vasudevan and Pranjal Awasthi and Adi Mayrav Gilady and Zelda Mariet and Ramy Eskander and Haiguang Li and Hexiang Hu and Guillermo Garrido and Philippe Schlattner and George Zhang and Rohun Saxena and Petar Dević and Kritika Muralidharan and Ashwin Murthy and Yiqian Zhou and Min Choi and Arissa Wongpanich and Zhengdong Wang and Premal Shah and Yuntao Xu and Yiling Huang and Stephen Spencer and Alice Chen and James Cohan and Junjie Wang and Jonathan Tompson and Junru Wu and Ruba Haroun and Haiqiong Li and Blanca Huergo and Fan Yang and Tongxin Yin and James Wendt and Michael Bendersky and Rahma Chaabouni and Javier Snaider and Johan Ferret and Abhishek Jindal and Tara Thompson and Andrew Xue and Will Bishop and Shubham Milind Phal and Archit Sharma and Yunhsuan Sung and Prabakar Radhakrishnan and Mo Shomrat and Reeve Ingle and Roopali Vij and Justin Gilmer and Mihai Dorin Istin and Sam Sobell and Yang Lu and Emily Nottage and Dorsa Sadigh and Jeremiah Willcock and Tingnan Zhang and Steve Xu and Sasha Brown and Katherine Lee and Gary Wang and Yun Zhu and Yi Tay and Cheolmin Kim and Audrey Gutierrez and Abhanshu Sharma and Yongqin Xian and Sungyong Seo and Claire Cui and Elena Pochernina and Cip Baetu and Krzysztof Jastrzębski and Mimi Ly and Mohamed Elhawaty and Dan Suh and Eren Sezener and Pidong Wang and Nancy Yuen and George Tucker and Jiahao Cai and Zuguang Yang and Cindy Wang and Alex Muzio and Hai Qian and Jae Yoo and Derek Lockhart and Kevin R. McKee and Mandy Guo and Malika Mehrotra and Artur Mendonça and Sanket Vaibhav Mehta and Sherry Ben and Chetan Tekur and Jiaqi Mu and Muye Zhu and Victoria Krakovna and Hongrae Lee and AJ Maschinot and Sébastien Cevey and HyunJeong Choe and Aijun Bai and Hansa Srinivasan and Derek Gasaway and Nick Young and Patrick Siegler and Dan Holtmann-Rice and Vihari Piratla and Kate Baumli and Roey Yogev and Alex Hofer and Hado van Hasselt and Svetlana Grant and Yuri Chervonyi and David Silver and Andrew Hogue and Ayushi Agarwal and Kathie Wang and Preeti Singh and Four Flynn and Josh Lipschultz and Robert David and Lizzetth Bellot and Yao-Yuan Yang and Long Le and Filippo Graziano and Kate Olszewska and Kevin Hui and Akanksha Maurya and Nikos Parotsidis and Weijie Chen and Tayo Oguntebi and Joe Kelley and Anirudh Baddepudi and Johannes Mauerer and Gregory Shaw and Alex Siegman and Lin Yang and Shravya Shetty and Subhrajit Roy and Yunting Song and Wojciech Stokowiec and Ryan Burnell and Omkar Savant and Robert Busa-Fekete and Jin Miao and Samrat Ghosh and Liam MacDermed and Phillip Lippe and Mikhail Dektiarev and Zach Behrman and Fabian Mentzer and Kelvin Nguyen and Meng Wei and Siddharth Verma and Chris Knutsen and Sudeep Dasari and Zhipeng Yan and Petr Mitrichev and Xingyu Wang and Virat Shejwalkar and Jacob Austin and Srinivas Sunkara and Navneet Potti and Yan Virin and Christian Wright and Gaël Liu and Oriana Riva and Etienne Pot and Greg Kochanski and Quoc Le and Gargi Balasubramaniam and Arka Dhar and Yuguo Liao and Adam Bloniarz and Divyansh Shukla and Elizabeth Cole and Jong Lee and Sheng Zhang and Sushant Kafle and Siddharth Vashishtha and Parsa Mahmoudieh and Grace Chen and Raphael Hoffmann and Pranesh Srinivasan and Agustin Dal Lago and Yoav Ben Shalom and Zi Wang and Michael Elabd and Anuj Sharma and Junhyuk Oh and Suraj Kothawade and Maigo Le and Marianne Monteiro and Shentao Yang and Kaiz Alarakyia and Robert Geirhos and Diana Mincu and Håvard Garnes and Hayato Kobayashi and Soroosh Mariooryad and Kacper Krasowiak and Zhixin and Lai and Shibl Mourad and Mingqiu Wang and Fan Bu and Ophir Aharoni and Guanjie Chen and Abhimanyu Goyal and Vadim Zubov and Ankur Bapna and Elahe Dabir and Nisarg Kothari and Kay Lamerigts and Nicola De Cao and Jeremy Shar and Christopher Yew and Nitish Kulkarni and Dre Mahaarachchi and Mandar Joshi and Zhenhai Zhu and Jared Lichtarge and Yichao Zhou and Hannah Muckenhirn and Vittorio Selo and Oriol Vinyals and Peter Chen and Anthony Brohan and Vaibhav Mehta and Sarah Cogan and Ruth Wang and Ty Geri and Wei-Jen Ko and Wei Chen and Fabio Viola and Keshav Shivam and Lisa Wang and Madeleine Clare Elish and Raluca Ada Popa and Sébastien Pereira and Jianqiao Liu and Raphael Koster and Donnie Kim and Gufeng Zhang and Sayna Ebrahimi and Partha Talukdar and Yanyan Zheng and Petra Poklukar and Ales Mikhalap and Dale Johnson and Anitha Vijayakumar and Mark Omernick and Matt Dibb and Ayush Dubey and Qiong Hu and Apurv Suman and Vaibhav Aggarwal and Ilya Kornakov and Fei Xia and Wing Lowe and Alexey Kolganov and Ted Xiao and Vitaly Nikolaev and Steven Hemingray and Bonnie Li and Joana Iljazi and Mikołaj Rybiński and Ballie Sandhu and Peggy Lu and Thang Luong and Rodolphe Jenatton and Vineetha Govindaraj and Hui and Li and Gabriel Dulac-Arnold and Wonpyo Park and Henry Wang and Abhinit Modi and Jean Pouget-Abadie and Kristina Greller and Rahul Gupta and Robert Berry and Prajit Ramachandran and Jinyu Xie and Liam McCafferty and Jianling Wang and Kilol Gupta and Hyeontaek Lim and Blaž Bratanič and Andy Brock and Ilia Akolzin and Jim Sproch and Dan Karliner and Duhyeon Kim and Adrian Goedeckemeyer and Noam Shazeer and Cordelia Schmid and Daniele Calandriello and Parul Bhatia and Krzysztof Choromanski and Ceslee Montgomery and Dheeru Dua and Ana Ramalho and Helen King and Yue Gao and Lynn Nguyen and David Lindner and Divya Pitta and Oleaser Johnson and Khalid Salama and Diego Ardila and Michael Han and Erin Farnese and Seth Odoom and Ziyue Wang and Xiangzhuo Ding and Norman Rink and Ray Smith and Harshal Tushar Lehri and Eden Cohen and Neera Vats and Tong He and Parthasarathy Gopavarapu and Adam Paszke and Miteyan Patel and Wouter Van Gansbeke and Lucia Loher and Luis Castro and Maria Voitovich and Tamara von Glehn and Nelson George and Simon Niklaus and Zach Eaton-Rosen and Nemanja Rakićević and Erik Jue and Sagi Perel and Carrie Zhang and Yuval Bahat and Angéline Pouget and Zhi Xing and Fantine Huot and Ashish Shenoy and Taylor Bos and Vincent Coriou and Bryan Richter and Natasha Noy and Yaqing Wang and Santiago Ontanon and Siyang Qin and Gleb Makarchuk and Demis Hassabis and Zhuowan Li and Mandar Sharma and Kumaran Venkatesan and Iurii Kemaev and Roxanne Daniel and Shiyu Huang and Saloni Shah and Octavio Ponce and Warren and Chen and Manaal Faruqui and Jialin Wu and Slavica Andačić and Szabolcs Payrits and Daniel McDuff and Tom Hume and Yuan Cao and MH Tessler and Qingze Wang and Yinan Wang and Ivor Rendulic and Eirikur Agustsson and Matthew Johnson and Tanya Lando and Andrew Howard and Sri Gayatri Sundara Padmanabhan and Mayank Daswani and Andrea Banino and Michael Kilgore and Jonathan Heek and Ziwei Ji and Alvaro Caceres and Conglong Li and Nora Kassner and Alexey Vlaskin and Zeyu Liu and Alex Grills and Yanhan Hou and Roykrong Sukkerd and Gowoon Cheon and Nishita Shetty and Larisa Markeeva and Piotr Stanczyk and Tejas Iyer and Yuan Gong and Shawn Gao and Keerthana Gopalakrishnan and Tim Blyth and Malcolm Reynolds and Avishkar Bhoopchand and Misha Bilenko and Dero Gharibian and Vicky Zayats and Aleksandra Faust and Abhinav Singh and Min Ma and Hongyang Jiao and Sudheendra Vijayanarasimhan and Lora Aroyo and Vikas Yadav and Sarah Chakera and Ashwin Kakarla and Vilobh Meshram and Karol Gregor and Gabriela Botea and Evan Senter and Dawei Jia and Geza Kovacs and Neha Sharma and Sebastien Baur and Kai Kang and Yifan He and Lin Zhuo and Marija Kostelac and Itay Laish and Songyou Peng and Louis O'Bryan and Daniel Kasenberg and Girish Ramchandra Rao and Edouard Leurent and Biao Zhang and Sage Stevens and Ana Salazar and Ye Zhang and Ivan Lobov and Jake Walker and Allen Porter and Morgan Redshaw and Han Ke and Abhishek Rao and Alex Lee and Hoi Lam and Michael Moffitt and Jaeyoun Kim and Siyuan Qiao and Terry Koo and Robert Dadashi and Xinying Song and Mukund Sundararajan and Peng Xu and Chizu Kawamoto and Yan Zhong and Clara Barbu and Apoorv Reddy and Mauro Verzetti and Leon Li and George Papamakarios and Hanna Klimczak-Plucińska and Mary Cassin and Koray Kavukcuoglu and Rigel Swavely and Alain Vaucher and Jeffrey Zhao and Ross Hemsley and Michael Tschannen and Heming Ge and Gaurav Menghani and Yang Yu and Natalie Ha and Wei He and Xiao Wu and Maggie Song and Rachel Sterneck and Stefan Zinke and Dan A. Calian and Annie Marsden and Alejandro Cruzado Ruiz and Matteo Hessel and Almog Gueta and Benjamin Lee and Brian Farris and Manish Gupta and Yunjie Li and Mohammad Saleh and Vedant Misra and Kefan Xiao and Piermaria Mendolicchio and Gavin Buttimore and Varvara Krayvanova and Nigamaa Nayakanti and Matthew Wiethoff and Yash Pande and Azalia Mirhoseini and Ni Lao and Jasmine Liu and Yiqing Hua and Angie Chen and Yury Malkov and Dmitry Kalashnikov and Shubham Gupta and Kartik Audhkhasi and Yuexiang Zhai and Sudhindra Kopalle and Prateek Jain and Eran Ofek and Clemens Meyer and Khuslen Baatarsukh and Hana Strejček and Jun Qian and James Freedman and Ricardo Figueira and Michal Sokolik and Olivier Bachem and Raymond Lin and Dia Kharrat and Chris Hidey and Pingmei Xu and Dennis Duan and Yin Li and Muge Ersoy and Richard Everett and Kevin Cen and Rebeca Santamaria-Fernandez and Amir Taubenfeld and Ian Mackinnon and Linda Deng and Polina Zablotskaia and Shashank Viswanadha and Shivanker Goel and Damion Yates and Yunxiao Deng and Peter Choy and Mingqing Chen and Abhishek Sinha and Alex Mossin and Yiming Wang and Arthur Szlam and Susan Hao and Paul Kishan Rubenstein and Metin Toksoz-Exley and Miranda Aperghis and Yin Zhong and Junwhan Ahn and Michael Isard and Olivier Lacombe and Florian Luisier and Chrysovalantis Anastasiou and Yogesh Kalley and Utsav Prabhu and Emma Dunleavy and Shaan Bijwadia and Justin Mao-Jones and Kelly Chen and Rama Pasumarthi and Emily Wood and Adil Dostmohamed and Nate Hurley and Jiri Simsa and Alicia Parrish and Mantas Pajarskas and Matt Harvey and Ondrej Skopek and Yony Kochinski and Javier Rey and Verena Rieser and Denny Zhou and Sun Jae Lee and Trilok Acharya and Guowang Li and Joe Jiang and Xiaofan Zhang and Bryant Gipson and Ethan Mahintorabi and Marco Gelmi and Nima Khajehnouri and Angel Yeh and Kayi Lee and Loic Matthey and Leslie Baker and Trang Pham and Han Fu and Alex Pak and Prakhar Gupta and Cristina Vasconcelos and Adam Sadovsky and Brian Walker and Sissie Hsiao and Patrik Zochbauer and Andreea Marzoca and Noam Velan and Junhao Zeng and Gilles Baechler and Danny Driess and Divya Jain and Yanping Huang and Lizzie Tao and John Maggs and Nir Levine and Jon Schneider and Erika Gemzer and Samuel Petit and Shan Han and Zach Fisher and Dustin Zelle and Courtney Biles and Eugene Ie and Asya Fadeeva and Casper Liu and Juliana Vicente Franco and Adrian Collister and Hao Zhang and Renshen Wang and Ruizhe Zhao and Leandro Kieliger and Kurt Shuster and Rui Zhu and Boqing Gong and Lawrence Chan and Ruoxi Sun and Sujoy Basu and Roland Zimmermann and Jamie Hayes and Abhishek Bapna and Jasper Snoek and Weel Yang and Puranjay Datta and Jad Al Abdallah and Kevin Kilgour and Lu Li and SQ Mah and Yennie Jun and Morgane Rivière and Abhijit Karmarkar and Tammo Spalink and Tao Huang and Lucas Gonzalez and Duc-Hieu Tran and Averi Nowak and John Palowitch and Martin Chadwick and Ellie Talius and Harsh Mehta and Thibault Sellam and Philipp Fränken and Massimo Nicosia and Kyle He and Aditya Kini and David Amos and Sugato Basu and Harrison Jobe and Eleni Shaw and Qiantong Xu and Colin Evans and Daisuke Ikeda and Chaochao Yan and Larry Jin and Lun Wang and Sachin Yadav and Ilia Labzovsky and Ramesh Sampath and Ada Ma and Candice Schumann and Aditya Siddhant and Rohin Shah and John Youssef and Rishabh Agarwal and Natalie Dabney and Alessio Tonioni and Moran Ambar and Jing Li and Isabelle Guyon and Benny Li and David Soergel and Boya Fang and Georgi Karadzhov and Cristian Udrescu and Trieu Trinh and Vikas Raunak and Seb Noury and Dee Guo and Sonal Gupta and Mara Finkelstein and Denis Petek and Lihao Liang and Greg Billock and Pei Sun and David Wood and Yiwen Song and Xiaobin Yu and Tatiana Matejovicova and Regev Cohen and Kalyan Andra and David D'Ambrosio and Zhiwei Deng and Vincent Nallatamby and Ebrahim Songhori and Rumen Dangovski and Andrew Lampinen and Pankil Botadra and Adam Hillier and Jiawei Cao and Nagabhushan Baddi and Adhi Kuncoro and Toshihiro Yoshino and Ankit Bhagatwala and Marcáurelio Ranzato and Rylan Schaeffer and Tianlin Liu and Shuai Ye and Obaid Sarvana and John Nham and Chenkai Kuang and Isabel Gao and Jinoo Baek and Shubham Mittal and Ayzaan Wahid and Anita Gergely and Bin Ni and Josh Feldman and Carrie Muir and Pascal Lamblin and Wolfgang Macherey and Ethan Dyer and Logan Kilpatrick and Víctor Campos and Mukul Bhutani and Stanislav Fort and Yanif Ahmad and Aliaksei Severyn and Kleopatra Chatziprimou and Oleksandr Ferludin and Mason Dimarco and Aditya Kusupati and Joe Heyward and Dan Bahir and Kevin Villela and Katie Millican and Dror Marcus and Sanaz Bahargam and Caglar Unlu and Nicholas Roth and Zichuan Wei and Siddharth Gopal and Deepanway Ghoshal and Edward Lee and Sharon Lin and Jennie Lees and Dayeong Lee and Anahita Hosseini and Connie Fan and Seth Neel and Marcus Wu and Yasemin Altun and Honglong Cai and Enrique Piqueras and Josh Woodward and Alessandro Bissacco and Salem Haykal and Mahyar Bordbar and Prasha Sundaram and Sarah Hodkinson and Daniel Toyama and George Polovets and Austin Myers and Anu Sinha and Tomer Levinboim and Kashyap Krishnakumar and Rachita Chhaparia and Tatiana Sholokhova and Nitesh Bharadwaj Gundavarapu and Ganesh Jawahar and Haroon Qureshi and Jieru Hu and Nikola Momchev and Matthew Rahtz and Renjie Wu and Aishwarya P S and Kedar Dhamdhere and Meiqi Guo and Umang Gupta and Ali Eslami and Mariano Schain and Michiel Blokzijl and David Welling and Dave Orr and Levent Bolelli and Nicolas Perez-Nieves and Mikhail Sirotenko and Aman Prasad and Arjun Kar and Borja De Balle Pigem and Tayfun Terzi and Gellért Weisz and Dipankar Ghosh and Aditi Mavalankar and Dhruv Madeka and Kaspar Daugaard and Hartwig Adam and Viraj Shah and Dana Berman and Maggie Tran and Steven Baker and Ewa Andrejczuk and Grishma Chole and Ganna Raboshchuk and Mahdi Mirzazadeh and Thais Kagohara and Shimu Wu and Christian Schallhart and Bernett Orlando and Chen Wang and Alban Rrustemi and Hao Xiong and Hao Liu and Arpi Vezer and Nolan Ramsden and Shuo-yiin Chang and Sidharth Mudgal and Yan Li and Nino Vieillard and Yedid Hoshen and Farooq Ahmad and Ambrose Slone and Amy Hua and Natan Potikha and Mirko Rossini and Jon Stritar and Sushant Prakash and Zifeng Wang and Xuanyi Dong and Alireza Nazari and Efrat Nehoran and Kaan Tekelioglu and Yinxiao Li and Kartikeya Badola and Tom Funkhouser and Yuanzhen Li and Varun Yerram and Ramya Ganeshan and Daniel Formoso and Karol Langner and Tian Shi and Huijian Li and Yumeya Yamamori and Amayika Panda and Alaa Saade and Angelo Scorza Scarpati and Chris Breaux and CJ Carey and Zongwei Zhou and Cho-Jui Hsieh and Sophie Bridgers and Alena Butryna and Nishesh Gupta and Vaibhav Tulsyan and Sanghyun Woo and Evgenii Eltyshev and Will Grathwohl and Chanel Parks and Seth Benjamin and Rina Panigrahy and Shenil Dodhia and Daniel De Freitas and Chris Sauer and Will Song and Ferran Alet and Jackson Tolins and Cosmin Paduraru and Xingyi Zhou and Brian Albert and Zizhao Zhang and Lei Shu and Mudit Bansal and Sarah Nguyen and Amir Globerson and Owen Xiao and James Manyika and Tom Hennigan and Rong Rong and Josip Matak and Anton Bakalov and Ankur Sharma and Danila Sinopalnikov and Andrew Pierson and Stephen Roller and Geoff Brown and Mingcen Gao and Toshiyuki Fukuzawa and Amin Ghafouri and Kenny Vassigh and Iain Barr and Zhicheng Wang and Anna Korsun and Rajesh Jayaram and Lijie Ren and Tim Zaman and Samira Khan and Yana Lunts and Dan Deutsch and Dave Uthus and Nitzan Katz and Masha Samsikova and Amr Khalifa and Nikhil Sethi and Jiao Sun and Luming Tang and Uri Alon and Xianghong Luo and Dian Yu and Abhishek Nayyar and Bryce Petrini and Will Truong and Vincent Hellendoorn and Nikolai Chinaev and Chris Alberti and Wei Wang and Jingcao Hu and Vahab Mirrokni and Ananth Balashankar and Avia Aharon and Aahil Mehta and Ahmet Iscen and Joseph Kready and Lucas Manning and Anhad Mohananey and Yuankai Chen and Anshuman Tripathi and Allen Wu and Igor Petrovski and Dawsen Hwang and Martin Baeuml and Shreyas Chandrakaladharan and Yuan Liu and Rey Coaguila and Maxwell Chen and Sally Ma and Pouya Tafti and Susheel Tatineni and Terry Spitz and Jiayu Ye and Paul Vicol and Mihaela Rosca and Adrià Puigdomènech and Zohar Yahav and Sanjay Ghemawat and Hanzhao Lin and Phoebe Kirk and Zaid Nabulsi and Sergey Brin and Bernd Bohnet and Ken Caluwaerts and Aditya Srikanth Veerubhotla and Dan Zheng and Zihang Dai and Petre Petrov and Yichong Xu and Ramin Mehran and Zhuo Xu and Luisa Zintgraf and Jiho Choi and Spurthi Amba Hombaiah and Romal Thoppilan and Sashank Reddi and Lukasz Lew and Li Li and Kellie Webster and KP Sawhney and Lampros Lamprou and Siamak Shakeri and Mayank Lunayach and Jianmin Chen and Sumit Bagri and Alex Salcianu and Ying Chen and Yani Donchev and Charlotte Magister and Signe Nørly and Vitor Rodrigues and Tomas Izo and Hila Noga and Joe Zou and Thomas Köppe and Wenxuan Zhou and Kenton Lee and Xiangzhu Long and Danielle Eisenbud and Anthony Chen and Connor Schenck and Chi Ming To and Peilin Zhong and Emanuel Taropa and Minh Truong and Omer Levy and Danilo Martins and Zhiyuan Zhang and Christopher Semturs and Kelvin Zhang and Alex Yakubovich and Pol Moreno and Lara McConnaughey and Di Lu and Sam Redmond and Lotte Weerts and Yonatan Bitton and Tiziana Refice and Nicolas Lacasse and Arthur Conmy and Corentin Tallec and Julian Odell and Hannah Forbes-Pollard and Arkadiusz Socala and Jonathan Hoech and Pushmeet Kohli and Alanna Walton and Rui Wang and Mikita Sazanovich and Kexin Zhu and Andrei Kapishnikov and Rich Galt and Matthew Denton and Ben Murdoch and Caitlin Sikora and Kareem Mohamed and Wei Wei and Uri First and Tim McConnell and Luis C. Cobo and James Qin and Thi Avrahami and Daniel Balle and Yu Watanabe and Annie Louis and Adam Kraft and Setareh Ariafar and Yiming Gu and Eugénie Rives and Charles Yoon and Andrei Rusu and James Cobon-Kerr and Chris Hahn and Jiaming Luo and Yuvein and Zhu and Niharika Ahuja and Rodrigo Benenson and Raphaël Lopez Kaufman and Honglin Yu and Lloyd Hightower and Junlin Zhang and Darren Ni and Lisa Anne Hendricks and Gabby Wang and Gal Yona and Lalit Jain and Pablo Barrio and Surya Bhupatiraju and Siva Velusamy and Allan Dafoe and Sebastian Riedel and Tara Thomas and Zhe Yuan and Mathias Bellaiche and Sheena Panthaplackel and Klemen Kloboves and Sarthak Jauhari and Canfer Akbulut and Todor Davchev and Evgeny Gladchenko and David Madras and Aleksandr Chuklin and Tyrone Hill and Quan Yuan and Mukundan Madhavan and Luke Leonhard and Dylan Scandinaro and Qihang Chen and Ning Niu and Arthur Douillard and Bogdan Damoc and Yasumasa Onoe and Fabian Pedregosa and Fred Bertsch and Chas Leichner and Joseph Pagadora and Jonathan Malmaud and Sameera Ponda and Andy Twigg and Oleksii Duzhyi and Jingwei Shen and Miaosen Wang and Roopal Garg and Jing Chen and Utku Evci and Jonathan Lee and Leon Liu and Koji Kojima and Masa Yamaguchi and Arunkumar Rajendran and AJ Piergiovanni and Vinodh Kumar Rajendran and Marco Fornoni and Gabriel Ibagon and Harry Ragan and Sadh MNM Khan and John Blitzer and Andrew Bunner and Guan Sun and Takahiro Kosakai and Scott Lundberg and Ndidi Elue and Kelvin Guu and SK Park and Jane Park and Arunachalam Narayanaswamy and Chengda Wu and Jayaram Mudigonda and Trevor Cohn and Hairong Mu and Ravi Kumar and Laura Graesser and Yichi Zhang and Richard Killam and Vincent Zhuang and Mai Giménez and Wael Al Jishi and Ruy Ley-Wild and Alex Zhai and Kazuki Osawa and Diego Cedillo and Jialu Liu and Mayank Upadhyay and Marcin Sieniek and Roshan Sharma and Tom Paine and Anelia Angelova and Sravanti Addepalli and Carolina Parada and Kingshuk Majumder and Avery Lamp and Sanjiv Kumar and Xiang Deng and Artiom Myaskovsky and Tea Sabolić and Jeffrey Dudek and Sarah York and Félix de Chaumont Quitry and Jiazhong Nie and Dee Cattle and Alok Gunjan and Bilal Piot and Waleed Khawaja and Seojin Bang and Simon Wang and Siavash Khodadadeh and Raghavender R and Praynaa Rawlani and Richard Powell and Kevin Lee and Johannes Griesser and GS Oh and Cesar Magalhaes and Yujia Li and Simon Tokumine and Hadas Natalie Vogel and Dennis Hsu and Arturo BC and Disha Jindal and Matan Cohen and Zi Yang and Junwei Yuan and Dario de Cesare and Tony Bruguier and Jun Xu and Monica Roy and Alon Jacovi and Dan Belov and Rahul Arya and Phoenix Meadowlark and Shlomi Cohen-Ganor and Wenting Ye and Patrick Morris-Suzuki and Praseem Banzal and Gan Song and Pranavaraj Ponnuramu and Fred Zhang and George Scrivener and Salah Zaiem and Alif Raditya Rochman and Kehang Han and Badih Ghazi and Kate Lee and Shahar Drath and Daniel Suo and Antonious Girgis and Pradeep Shenoy and Duy Nguyen and Douglas Eck and Somit Gupta and Le Yan and Joao Carreira and Anmol Gulati and Ruoxin Sang and Daniil Mirylenka and Emma Cooney and Edward Chou and Mingyang Ling and Cindy Fan and Ben Coleman and Guilherme Tubone and Ravin Kumar and Jason Baldridge and Felix Hernandez-Campos and Angeliki Lazaridou and James Besley and Itay Yona and Neslihan Bulut and Quentin Wellens and AJ Pierigiovanni and Jasmine George and Richard Green and Pu Han and Connie Tao and Geoff Clark and Chong You and Abbas Abdolmaleki and Justin Fu and Tongzhou Chen and Ashwin Chaugule and Angad Chandorkar and Altaf Rahman and Will Thompson and Penporn Koanantakool and Mike Bernico and Jie Ren and Andrey Vlasov and Sergei Vassilvitskii and Maciej Kula and Yizhong Liang and Dahun Kim and Yangsibo Huang and Chengxi Ye and Dmitry Lepikhin and Wesley Helmholz},
      year={2025},
      eprint={2507.06261},
      archivePrefix={arXiv},
      primaryClass={cs.CL},
      url={https://arxiv.org/abs/2507.06261}, 
}

@inproceedings{TheC3,
  title={The Claude 3 Model Family: Opus, Sonnet, Haiku},
  author={Claude},
  url={https://api.semanticscholar.org/CorpusID:268232499}
}

@misc{adelani2021masakhanernamedentityrecognition,
      title={MasakhaNER: Named Entity Recognition for African Languages}, 
      author={David Ifeoluwa Adelani and Jade Abbott and Graham Neubig and Daniel D'souza and Julia Kreutzer and Constantine Lignos and Chester Palen-Michel and Happy Buzaaba and Shruti Rijhwani and Sebastian Ruder and Stephen Mayhew and Israel Abebe Azime and Shamsuddeen Muhammad and Chris Chinenye Emezue and Joyce Nakatumba-Nabende and Perez Ogayo and Anuoluwapo Aremu and Catherine Gitau and Derguene Mbaye and Jesujoba Alabi and Seid Muhie Yimam and Tajuddeen Gwadabe and Ignatius Ezeani and Rubungo Andre Niyongabo and Jonathan Mukiibi and Verrah Otiende and Iroro Orife and Davis David and Samba Ngom and Tosin Adewumi and Paul Rayson and Mofetoluwa Adeyemi and Gerald Muriuki and Emmanuel Anebi and Chiamaka Chukwuneke and Nkiruka Odu and Eric Peter Wairagala and Samuel Oyerinde and Clemencia Siro and Tobius Saul Bateesa and Temilola Oloyede and Yvonne Wambui and Victor Akinode and Deborah Nabagereka and Maurice Katusiime and Ayodele Awokoya and Mouhamadane MBOUP and Dibora Gebreyohannes and Henok Tilaye and Kelechi Nwaike and Degaga Wolde and Abdoulaye Faye and Blessing Sibanda and Orevaoghene Ahia and Bonaventure F. P. Dossou and Kelechi Ogueji and Thierno Ibrahima DIOP and Abdoulaye Diallo and Adewale Akinfaderin and Tendai Marengereke and Salomey Osei},
      year={2021},
      eprint={2103.11811},
      archivePrefix={arXiv},
      primaryClass={cs.CL},
      url={https://arxiv.org/abs/2103.11811}, 
}

@inproceedings{kakwani-etal-2020-indicnlpsuite,
    title = "{I}ndic{NLPS}uite: Monolingual Corpora, Evaluation Benchmarks and Pre-trained Multilingual Language Models for {I}ndian Languages",
    author = "Kakwani, Divyanshu  and
      Kunchukuttan, Anoop  and
      Golla, Satish  and
      N.C., Gokul  and
      Bhattacharyya, Avik  and
      Khapra, Mitesh M.  and
      Kumar, Pratyush",
    editor = "Cohn, Trevor  and
      He, Yulan  and
      Liu, Yang",
    booktitle = "Findings of the Association for Computational Linguistics: EMNLP 2020",
    month = nov,
    year = "2020",
    address = "Online",
    publisher = "Association for Computational Linguistics",
    url = "https://aclanthology.org/2020.findings-emnlp.445/",
    doi = "10.18653/v1/2020.findings-emnlp.445",
    pages = "4948--4961"
}

@inproceedings{FLORES,
  title={The FLORES-101  Evaluation Benchmark for Low-Resource and Multilingual Machine Translation},
  author={Goyal, Naman and Gao, Cynthia and Chaudhary, Vishrav and Chen, Peng-Jen and Wenzek, Guillaume and Ju, Da and Krishnan, Sanjana and Ranzato, Marc'Aurelio and Guzm\'{a}n, Francisco and Fan, Angela},
  year={2021}
}

@inproceedings{yang-etal-2022-cino,
    title = "{CINO}: A {C}hinese Minority Pre-trained Language Model",
    author = "Yang, Ziqing  and
      Xu, Zihang  and
      Cui, Yiming  and
      Wang, Baoxin  and
      Lin, Min  and
      Wu, Dayong  and
      Chen, Zhigang",
    booktitle = "Proceedings of the 29th International Conference on Computational Linguistics",
    month = oct,
    year = "2022",
    address = "Gyeongju, Republic of Korea",
    publisher = "International Committee on Computational Linguistics",
    url = "https://aclanthology.org/2022.coling-1.346",
    pages = "3937--3949"
}

\appendix


\section{Detailed Dataset Statistics}
\label{sec:appendix.detail_stats}

\subsection{Exact subset sizes}
LaoBench consists of three subsets:
\begin{itemize}[leftmargin=*]
\vspace{-6pt}
    \item \textbf{Lao-7k}: \textbf{7,000} open-source multiple-choice questions.
    \vspace{-6pt}
    \item \textbf{Lao-10k}: \textbf{10k} closed-source multiple-choice questions used for secure black-box evaluation.
    \vspace{-6pt}
    \item \textbf{Lao-500}: \textbf{500} open-ended prompts for generation evaluation.
\end{itemize}

\subsection{Category and subdomain distribution}
Table~\ref{tab:appendix_distribution} reports the distribution of instances across categories and subdomains. 
We ensure balanced coverage by enforcing minimum instance counts per subdomain during construction.

\begin{table}[!h]
\centering
\scriptsize
\setlength{\tabcolsep}{7pt}
\renewcommand{\arraystretch}{1.1}

\begin{tabular}{llr}
\toprule
\rowcolor{mheader}
\textbf{Category} & \textbf{Subdomain} & \textbf{\# Instances} \\
\midrule

\rowcolor{mgroup}
\multicolumn{3}{l}{\textbf{Knowledge Application}} \\
History \& Development &  & 425 \\
\rowcolor{malt}
Politics \& Law & & 695 \\
Society \& Culture & & 566 \\
\rowcolor{malt}
Nature \& Science & & 314 \\

\midrule
\rowcolor{mgroup}
\multicolumn{3}{l}{\textbf{K12 Education}} \\
Humanities \& Arts & & 305 \\
\rowcolor{malt}
Health \& Environment & & 213 \\
Thinking \& Philosophy & & 764 \\
\rowcolor{malt}
Social Sciences & & 825 \\
Natural Sciences & & 893 \\

\midrule
\rowcolor{mgroup}
\multicolumn{3}{l}{\textbf{Translation}} \\
International Affairs & & 276 \\
\rowcolor{malt}
Culture \& History & & 483 \\
Environment \& Development & & 388 \\
\rowcolor{malt}
Society \& Law & & 853 \\

\bottomrule
\end{tabular}

\caption{Detailed distribution of LaoBench instances across categories and subdomains.}
\label{tab:appendix_distribution}
\vspace{-4pt}
\end{table}

\paragraph{Fine-grained taxonomy.}
Beyond the coarse-grained subdomain categorization, we further define a three-level hierarchical taxonomy 
(\textit{dimension} $\rightarrow$ \textit{category} $\rightarrow$ \textit{subcategory}) 
to ensure consistent data construction and coverage control.
Table~\ref{tab:taxonomy} provides the full taxonomy used in LaoBench, including all third-level subcategories.
\begin{table*}[htbp]

\centering
\small
\setlength{\tabcolsep}{6pt}
\renewcommand{\arraystretch}{1.12}

\begin{tabular}{p{0.20\linewidth} p{0.28\linewidth} p{0.45\linewidth}}
\toprule
\rowcolor{mheader}
\textbf{Dimension} & \textbf{Category} & \textbf{Subcategory} \\
\midrule\multirow{18}{*}{Knowledge Application} 
& History and Development & Historical Figures and Events \\
&                         & Wars and Conflicts \\
&                         & Political Changes and Development \\
&                         & Education and Progress \\
\cmidrule(lr){2-3}
& Politics and Law        & Peace Agreements and Civil War Resolution \\
&                         & Diplomacy and International Relations \\
&                         & Political and Administrative Systems \\
&                         & Laws and Conventions \\
&                         & Borders and Territorial Demarcation \\
\cmidrule(lr){2-3}
& Society and Culture     & Religion and Beliefs \\
&                         & Culture and Traditions \\
&                         & Ethnic Migration and Social Development \\
&                         & Commemorative Days and Cultural Heritage \\
\cmidrule(lr){2-3}
& Nature and Science      & Geography and Environment \\
&                         & Archaeological Discoveries and Anthropology \\
\midrule
\multirow{20}{*}{K12 Foundational Education} 
& Humanities and Arts     & Writing and Media \\
&                         & Behavioral Norms and Etiquette \\
\cmidrule(lr){2-3}
& Health and Environment  & Health and Psychology \\
&                         & Environmental Protection and Sustainability \\
\cmidrule(lr){2-3}
& Thinking and Philosophy & Mathematics and Logical Thinking \\
&                         & Moral Philosophy and Common Sense \\
\cmidrule(lr){2-3}
& Social Sciences         & History and Geography \\
&                         & Historical Events and Culture \\
&                         & Local Culture and History \\
&                         & Social Sciences and Political Science \\
&                         & Social Responsibility and Civic Awareness \\
\cmidrule(lr){2-3}
& Natural Sciences        & Zoology and Botany \\
&                         & Physics and Technological Inventions \\
&                         & Matter and Chemistry \\
&                         & Fundamentals of Biology \\
\midrule
\multirow{15}{*}{Translation} 
& International Affairs   & International Relations and Treaties \\
&                         & Impact of Border Agreements \\
\cmidrule(lr){2-3}
& Culture and History     & Religion and Festival Customs \\
&                         & Lao History and Culture \\
&                         & Archaeology and Historical Research \\
&                         & Natural Landscapes and Cultural Heritage \\
\cmidrule(lr){2-3}
& Environment and Development & Education System and Social Equity \\
&                         & Natural Resources and Environmental Protection \\
\cmidrule(lr){2-3}
& Society and Law         & Civil Rights and Development History \\
&                         & Laws, Regulations, and Education \\
&                         & Society and Mental Health \\
&                         & Economic Development and Legal Systems \\
&                         & Ethics and Legal Responsibility \\
\bottomrule
\end{tabular}
\caption{Full hierarchical taxonomy of LaoBench (dimension $\rightarrow$ category $\rightarrow$ subcategory), used to guide data construction and ensure balanced coverage across fine-grained topics.}
\label{tab:taxonomy}
\end{table*}

\section{Data Construction and Validation Protocols}
\label{sec:appendix.qa}

\subsection{Data sources}
We curate raw materials from authoritative Lao sources, including:
(i) national K12 textbooks and curriculum guidelines,
(ii) government and legal documents,
(iii) encyclopedic and educational publications,
(iv) culturally grounded articles and local knowledge resources.
All sources are documented and archived internally. We release metadata summaries and source-type statistics for the open-source subset.

\subsection{Expert annotation roles and training}
Our annotation team includes Lao linguists, education experts, and subject-matter specialists.
Before annotation, we conduct a standardized training session covering:
question writing standards, distractor design principles, difficulty calibration, and sensitivity guidelines.

\paragraph{Annotator statistics.}
We employ 11 Lao-native annotators and 4 expert reviewers.
Each item is authored by one annotator and reviewed by at least 2 independent experts.
Disagreements are resolved through adjudication by a senior reviewer.

\subsection{Multiple-choice question construction guidelines}
For each MCQ:
\begin{itemize}[leftmargin=*]
\vspace{-6pt}
    \item The stem is written in natural Lao and references grounded Lao contexts.
    \vspace{-6pt}
    \item Four options are provided with exactly one correct answer.
    \vspace{-6pt}
    \item Distractors are designed to be plausible and semantically close to the correct option, avoiding trivial elimination.
    \vspace{-6pt}
    \item We prohibit “all of the above” and “none of the above” to reduce ambiguity.
    \vspace{-6pt}
    \item We remove items whose correct option depends on missing context or relies on overly specific memorization.
\end{itemize}

\subsection{Translation instance construction}
Each translation instance is constructed by:
(i) selecting representative sentences from authoritative sources,
(ii) producing a reference translation by professional translators,
(iii) verifying semantic alignment and Lao fluency via expert review.
We cover translation directions among Lao, Chinese, and English according to the benchmark taxonomy.

\subsection{Agent-assisted verification and filtering}
\label{sec:appendix.agent_verification}
We employ agent-based verification to support scalability and consistency.
Agents are used for:
\begin{itemize}[leftmargin=*]
\vspace{-6pt}
    \item \textbf{Duplicate / near-duplicate detection:} lexical overlap (character n-grams) and semantic similarity (embedding retrieval).
    \vspace{-6pt}
    \item \textbf{Semantic consistency:} verifying that the correct option is uniquely supported by the stem and that distractors are not also correct.
    \vspace{-6pt}
    \item \textbf{Context independence:} removing items requiring external context not provided in the prompt.
    \vspace{-6pt}
    \item \textbf{Sensitivity screening:} detecting potentially harmful, private, or politically sensitive content, followed by expert review.
\end{itemize}

\subsection{Inter-annotator agreement and adjudication}
To estimate reliability, we sample 500 items and ask 3 annotators to independently label the correct answer.
We report Fleiss’ $\kappa$:
\[
\kappa = \frac{p_o - p_e}{1 - p_e},
\]
where $p_o$ is observed agreement and $p_e$ is chance agreement.
Our measured agreement is $\kappa = 0.87$, indicating substantial consistency.

\subsection{Expert Contributor Profiles and Transparency}
\label{sec:appendix.contributor_profiles}

To ensure the linguistic integrity and cultural authenticity of LaoBench, we engaged a diverse group of 55 human contributors throughout the construction, validation, and auditing phases. Their roles, academic backgrounds, and specific responsibilities are summarized in Table~\ref{tab:contributor_stats}.

All culturally grounded and curriculum-aligned items underwent a double-blind review process by at least two qualified contributors. Items identified with semantic ambiguity, insufficient contextual grounding, or potential cultural insensitivity were either significantly revised or discarded. Final dataset audits were conducted by senior expert reviewers (predominantly PhD-level faculty) to guarantee alignment with Lao national educational standards.

\begin{table*}[ht]
\centering
\small
\renewcommand{\arraystretch}{1.3}
\setlength{\tabcolsep}{8pt}
\begin{tabularx}{\textwidth}{l p{4cm} X l c}
\toprule
\rowcolor{mheader}
\textbf{Role} & \textbf{Background} & \textbf{Task Description} & \textbf{Education Level (Count)} & \textbf{Total} \\
\midrule

Domain Experts & 
Lao universities \& educational institutions & 
Culturally grounded knowledge \& K12 dataset creation & 
Bachelor (16), Master (5), PhD (4) & 
25 \\

\rowcolor{malt}
Linguistic Experts & 
Lao–Chinese / Lao–English translation scholars & 
Parallel corpus construction \& random spot-check auditing & 
Bachelor (6), Master (3), PhD (2) & 
11 \\

Senior Reviewers & 
Lao universities \& Dept. of Education & 
Final comprehensive full audit across all categories & 
Master (2), PhD (8) & 
10 \\

\rowcolor{malt}
Data Curators & 
Computer Science \& NLP backgrounds & 
Data consolidation, formatting, \& consistency verification & 
Bachelor (5), Master (2), PhD (2) & 
9 \\

\midrule
\textbf{Total} & & & & \textbf{55} \\
\bottomrule
\end{tabularx}
\caption{Summary of human contributors involved in the LaoBench pipeline. The team composition ensures a balance between pedagogical expertise, linguistic precision, and technical data integrity.}
\label{tab:contributor_stats}
\end{table*}

\section{Translation Evaluation Configuration}
\label{sec:appendix.translation_eval}

\subsection{Tokenization and segmentation}
We evaluate translation outputs under a standardized reference-based protocol.
Since Lao is written in a \textit{scriptio continua} style without explicit whitespace word boundaries,
both BLEU and chrF++ can be sensitive to segmentation.
We therefore perform Lao-aware word segmentation using LaoNLP (v0.7) on both system outputs and references prior to metric computation.
For non-Lao languages (Chinese and English), we apply standard whitespace tokenization (English) and character-aware tokenization (Chinese) as provided by SacreBLEU.

\subsection{BLEU}
We compute corpus-level BLEU using SacreBLEU with a unified configuration across all systems.
We report BLEU scores on the tokenized text, which improves stability for Lao evaluation.

\subsection{ChrF++}
In addition to BLEU, we report \textbf{chrF++} to capture character n-gram overlap and reduce sensitivity to tokenization.
chrF++ computes a character-level F-score with word boundary awareness, which is particularly suitable for low-resource and scriptio continua languages such as Lao.
We compute chrF++ using SacreBLEU with the default setting (character n-gram order up to 6 and word n-gram order up to 2), and report corpus-level scores for each translation subdomain.

\begin{table}[!t]
\centering
\scriptsize
\setlength{\tabcolsep}{3pt}
\renewcommand{\arraystretch}{1.15}
\rowcolors{2}{mgray}{white}
\begin{tabular}{lcccc}
\toprule
\rowcolor{mheader}
\textbf{Model} & \textbf{Soc.\& Law} & \textbf{Cult.\& Hist.} & \textbf{Inter.\& Aff.} & \textbf{Env.\& Dev.} \\
\midrule
GPT-5-High     & 50.31 & 59.92 & 66.08 & 57.41 \\
Gemini-2.5-Pro & 52.84 & 56.62 & 66.35 & 60.12 \\
Qwen3-Max      & 52.38 & 58.04 & 65.21 & 58.93 \\
\bottomrule
\end{tabular}
\caption{chrF++ scores on LaoBench translation subsets. Higher is better.}
\label{tab:translation_chrf}
\vspace{-6pt}
\end{table}

\section{Arena-Style Evaluation Reliability}
\label{sec:appendix.arena_reliability}
We enforce a strict JSON-only output format for judges to ensure deterministic parsing and prevent explanation leakage.

\subsection{Judge agreement}
We compute rank correlation between judges (Gemini-2.5-Pro and Qwen3-Max) using Spearman’s $\rho$ and Kendall’s $\tau$.
Our measured agreement is:
\begin{itemize}[leftmargin=*]
\vspace{-6pt}
    \item Spearman $\rho$: 0.83
    \vspace{-6pt}
    \item Kendall $\tau$: 0.71
\end{itemize}

\subsection{Human sanity check}
We randomly sample 50 prompts from Lao-500 and ask Lao-native human evaluators to judge pairwise outputs.
We compare human preferences with LLM judges and observe 84\% agreement, supporting judge reliability.

\subsection{Baseline anchoring analysis}
To quantify baseline effects, we repeat evaluation using an alternative baseline model (Claude-Opus-4.1) and measure ranking stability.
We find that while absolute win-rates shift, relative rankings remain largely stable.

\section{Arena Prompt Templates and Output Parsing}
\label{sec:appendix.arena_prompts}

To ensure transparency and reproducibility, we provide the exact prompt templates used for Arena-style pairwise evaluation on Lao-500.
Each comparison consists of a user prompt and two candidate answers (one from the baseline model and one from a challenger model),
evaluated by an independent judge model.
We enforce a strict JSON-only output format to enable deterministic parsing and to avoid explanation leakage.
To mitigate position bias, we randomize answer ordering and run each comparison twice with swapped positions, then average the outcomes.

\subsection{Generation and Judge Prompt Template}
\label{sec:appendix.gen_prompt}

For each Lao-500 prompt, we generate one response from the baseline model and one response from a challenger model using the same generation template.
To reflect realistic user-facing usage in Laos and to avoid code-switching into English or Chinese, we explicitly enforce Lao-only outputs.
All models are evaluated in a zero-shot setting with deterministic decoding (temperature set to 0 when supported).

\paragraph{Generation prompt template.}
Table~\ref{tab:arena-generation-prompt} shows the prompt template used to generate model responses.
The template consists of (i) a system instruction enforcing Lao-only outputs, and (ii) the user prompt drawn from Lao-500.
For models that do not support system prompts, the system instruction is prepended to the user message.

\begin{table*}[t!]
\centering
\small
\renewcommand{\arraystretch}{1.55}
\setlength{\tabcolsep}{8pt}

\begin{tabularx}{\textwidth}{l X}
\noalign{\hrule height 1.2pt}
\rowcolor{MorandiBlue}
\textbf{\color{white}Type} &
\textbf{\color{white}Generation Prompt Template (Visualized Layout)} \\
\noalign{\hrule height 0.8pt}

\rowcolor{MorandiBlueLight}
\textbf{System} &
You are a helpful assistant for Lao-speaking users.
Answer the user prompt in \textbf{fluent and natural Lao}.
Do \textbf{not} switch to English or Chinese unless explicitly requested.\par\par

\textbf{Rules:}
\begin{itemize}[leftmargin=*, itemsep=1pt, topsep=2pt]
    \item Respond only in Lao.
    \item Prioritize correctness and clarity.
    \item Avoid unnecessary verbosity; be concise but complete.
\end{itemize}
\\
\hline

\rowcolor{MorandiGreenLight}
\textbf{User} &
\textbf{[User Prompt]} \\
\noalign{\hrule height 1.2pt}
\end{tabularx}

\caption{
Visualized generation prompt template used to obtain candidate responses on Lao-500.
Both the baseline model and each challenger model are prompted with the same template to ensure fair comparison.
}
\label{tab:arena-generation-prompt}
\vspace{-6pt}
\end{table*}

\paragraph{Judge prompt template.}
Table~\ref{tab:arena-judge-prompt} shows the judge prompt used for pairwise evaluation.
Given the same user prompt, the judge receives two anonymized candidate responses (Response A and Response B) and decides which is better.
The judge is instructed to evaluate correctness, completeness, reasoning quality, clarity, and Lao fluency.
To mitigate position bias, we evaluate each pair twice by swapping the positions of A and B and then average the outcomes.

\label{sec:appendix.judge_prompt}

\begin{table*}[t!]
\centering
\small
\renewcommand{\arraystretch}{1.55}
\setlength{\tabcolsep}{8pt}

\begin{tabularx}{\textwidth}{l X}
\noalign{\hrule height 1.2pt}
\rowcolor{MorandiBlue}
\textbf{\color{white}Type} &
\textbf{\color{white}Arena Judge Prompt Template (Visualized Layout)} \\
\noalign{\hrule height 0.8pt}

\rowcolor{MorandiBlueLight}
\textbf{System} &
You are a strict and fair evaluator for \textbf{Lao-language} answers. 
You will be shown a user prompt and two candidate answers (Answer A and Answer B). 
Your task is to decide which answer is better for a Lao-speaking user.\par\par

\textbf{Evaluation Criteria:}
(1) Correctness,
(2) Completeness,
(3) Reasoning Quality,
(4) Clarity \& Structure,
(5) Lao Fluency \& Appropriateness.\par\par

\textbf{Rules:}\par
\begin{itemize}[leftmargin=*, itemsep=1pt, topsep=2pt]
    \item Do \textbf{not} favor verbosity. Prefer concise but complete answers.
    \item If both answers are similarly good or similarly bad, output \textbf{Tie}.
    \item The answer order is randomized. Do not assume A is better than B.
    \item Do \textbf{not} reveal your reasoning or analysis.
    \item Output must follow the required JSON format exactly.
\end{itemize}

\textbf{Output Format (must be exact):}\par
\texttt{\{"winner": "A"\}} \quad or \quad 
\texttt{\{"winner": "B"\}} \quad or \quad 
\texttt{\{"winner": "Tie"\}} \\

\hline

\rowcolor{MorandiGreenLight}
\textbf{User} &
\textbf{[User Prompt]} \par\par
\textbf{Answer A:}\par
\textbf{[Answer A]} \par\par
\textbf{Answer B:}\par
\textbf{[Answer B]} \\

\noalign{\hrule height 1.2pt}
\end{tabularx}

\caption{Visualized Arena judge prompt template used in Lao-500 evaluation. The judge outputs only a JSON decision (\texttt{A/B/Tie}) to support deterministic parsing and reduce judge bias.}
\label{tab:arena-judge-prompt}
\vspace{-6pt}
\end{table*}

\subsection{Tie Handling and Scoring}
A tie is assigned a half-win (0.5) for each model. 
We report the final win rate of each challenger against the baseline across all prompts.

\subsection{Position Bias Mitigation}
For each prompt, we run the judge twice with swapped ordering (A/B). 
We compute the bias-corrected win signal $w_i^{J}(M)$ by averaging the two outcomes.

\subsection{Output Parsing Rules}
We parse the judge output as a JSON object with a single field \texttt{winner}, 
whose value must be exactly one of \texttt{"A"}, \texttt{"B"}, or \texttt{"Tie"}.
Outputs that fail JSON parsing or contain invalid values are re-queried once.
If the output remains invalid, we assign a tie to avoid introducing systematic bias.

\section{Judge-Specific Arena Results and Cross-Judge Gap}
\label{sec:appendix.arena_judge_breakdown}

Table~\ref{tab:arena_judge_breakdown} reports judge-specific win rates and 95\% bootstrap confidence intervals for each model on Lao-500.
We also report $\Delta$(G$-$Q), the difference between Gemini-2.5-Pro and Qwen3-Max judges, and \textbf{Gap}=$|\Delta|$ to quantify judge sensitivity.
We observe that some model families show systematic preference shifts under different judges, motivating our multi-judge aggregation protocol in the main paper.
We further verify judge consistency by reporting cross-judge rank correlation (Spearman $\rho$ and Kendall $\tau$) in Appendix~\ref{sec:appendix.arena_reliability}.

\begin{table*}[t]
\centering
\small
\setlength{\tabcolsep}{6pt}
\renewcommand{\arraystretch}{1.15}
\altrowcolors
\begin{tabular}{lcccccc}
\toprule
\rowcolor{mheader}
\textbf{Model} & \textbf{Gemini} & \textbf{Gemini 95\% CI} & \textbf{Qwen} & \textbf{Qwen 95\% CI} & {$\Delta$(G$-$Q)} & \textbf{Gap} \\
\midrule
Gemini-2.5-Pro                      
& 54.22 & [52.51, 55.93] & 48.85 & [46.60, 51.10] & +5.37 & \cellcolor{mred}5.37 \\
Claude-Sonnet-4.5-20250929-thinking 
& 50.50 & [48.35, 52.65] & 50.08 & [47.51, 52.65] & +0.42 & 0.42 \\
GPT-5-High (baseline)               
& 49.94 & [48.00, 51.88] & 49.94 & [47.99, 51.89] & +0.00 & 0.00 \\
Claude-Opus-4.1-20250805            
& 50.96 & [48.91, 53.01] & 47.64 & [45.26, 50.02] & +3.32 & 3.32 \\
Qwen3-Max                           
& 45.16 & [43.16, 47.16] & 52.80 & [49.86, 55.74] & -7.64 & \cellcolor{mred}7.64 \\
Qwen3-235B-A22B-Instruct-2507       
& 45.53 & [43.50, 47.56] & 51.75 & [48.95, 54.55] & -6.22 & \cellcolor{mred}6.22 \\
DeepSeek-V3.2-Exp(Thinking)         
& 48.69 & [46.48, 50.90] & 47.29 & [44.67, 49.91] & +1.40 & 1.40 \\
Qwen3-Plus(Non-Thinking)            
& 45.91 & [43.91, 47.91] & 49.35 & [46.19, 52.51] & -3.44 & 3.44 \\
DeepSeek-V3.2-Exp(Non-Thinking)     
& 47.96 & [45.41, 50.51] & 46.36 & [43.07, 49.65] & +1.60 & 1.60 \\
Qwen3-Plus(Thinking)                
& 47.74 & [45.15, 50.33] & 45.20 & [43.31, 47.09] & +2.54 & 2.54 \\
Qwen3-235B-A22B                     
& 46.31 & [44.15, 48.47] & 45.53 & [42.65, 48.41] & +0.78 & 0.78 \\
Qwen3-Next-80B-A3B-Instruct         
& 44.55 & [41.85, 47.25] & 45.73 & [42.48, 48.98] & -1.18 & 1.18 \\
\bottomrule
\end{tabular}
\caption{
Judge-specific Arena win rates (\%) on Lao-500 with 95\% bootstrap confidence intervals (CI).
$\Delta$(G$-$Q) denotes the score difference between Gemini-2.5-Pro and Qwen3-Max judges, and \textbf{Gap} is the absolute difference.
}
\label{tab:arena_judge_breakdown}
\end{table*}

\section{Contamination and Overlap Checks}
\label{sec:appendix.contamination}
We perform two forms of contamination analysis on the open-source subset (Lao-7k):
\begin{itemize}[leftmargin=*]
\vspace{-6pt}
    \item \textbf{Web overlap retrieval:} searching question stems and key phrases via web search and checking for exact matches.
    \vspace{-6pt}
    \item \textbf{N-gram overlap:} measuring overlap between benchmark text and public corpora (e.g., Lao Wikipedia, news dumps).
\end{itemize}
We observe that 6.2\% of items have potential overlap candidates. Manual inspection suggests that most cases are attributable to common factual statements rather than direct test leakage.

\section{Black-box Evaluation Protocol for Lao-10k}
\label{sec:appendix.blackbox_protocol}
\paragraph{Black-box evaluation protocol.}
To enable secure and contamination-resistant evaluation, we keep Lao-10k hidden and evaluate models through a black-box protocol.
Participants submit either (i) predicted option labels for a provided set of item IDs, or (ii) an inference API endpoint that follows our standardized prompt template.
We return only aggregated scores (overall and subdomain-level accuracies) without per-item feedback, and enforce rate limits on submissions to reduce leaderboard overfitting.
We will deploy this protocol as an online evaluation service upon publication.

\section{Error Analysis}
\label{sec:appendix.error_analysis}

\subsection{Knowledge Application failure modes}
We categorize model errors into:
\begin{enumerate}[leftmargin=*]
\vspace{-6pt}
    \item \textbf{Cultural grounding errors:} misunderstanding Lao-specific conventions or institutions.
    \vspace{-6pt}
    \item \textbf{Reasoning errors:} failing multi-step inference even with correct knowledge.
    \vspace{-6pt}
    \item \textbf{Lexical confusion:} confusion caused by loanwords, named entities, or polysemy.
\end{enumerate}

\subsection{Translation error types}
We analyze translation errors and identify:
(i) terminology mistranslation,
(ii) omission or hallucination,
(iii) incorrect formal register,
(iv) word-order and fluency degradation.
We find that culturally grounded and legal/administrative domains exhibit the highest error rates.

\section{Human Performance Protocol}
\label{sec:appendix.human_performance}
Human expert performance is measured by assigning 3 Lao-native experts to answer MCQ items without external tools.
Each item is answered independently. We report mean accuracy and standard deviation across experts.
Human accuracy exceeds 97\% across all subdomains, confirming benchmark validity and meaningful headroom for future research.

\end{document}